  \documentclass{article}

  \usepackage{graphicx}
  \usepackage{subcaption}
  \usepackage{wrapfig}
  \usepackage{booktabs} 
  \usepackage[normalem]{ulem}

  \usepackage{hyperref}

  \usepackage[accepted]{icml2026/icml2026}

  \usepackage{amsmath}
  \usepackage{amssymb}
  \usepackage{mathtools}
  \usepackage{amsthm}
  \usepackage{bbm}       
  \usepackage{placeins}  

  \usepackage{graphicx}
  \usepackage{tikz-cd}
  \usepackage{tikz}
  \usetikzlibrary{shapes.geometric, arrows.meta, positioning, shapes.symbols, calc}
  \usepackage{multicol}
  \usepackage{booktabs}

  \usepackage[capitalize,noabbrev]{cleveref}

  \usepackage[textsize=tiny]{todonotes}
  \usepackage{multirow}
  \usepackage{xfrac}

\expandafter\let\csname algorithm*\endcsname\relax
\expandafter\let\csname endalgorithm*\endcsname\relax

\usepackage[ruled,lined,linesnumbered]{algorithm2e}
\usepackage{booktabs}
  \usepackage{soul}
  \definecolor{LLdark}{RGB}{0,51,153}      
  \definecolor{LLlight}{RGB}{204,229,255}  
  \definecolor{KKdark}{RGB}{153,0,0}       
  \definecolor{KKlight}{RGB}{255,204,204}  

  \icmlsetsymbol{core}{*}
  \icmlsetsymbol{lead}{$\dagger$}

\begin{document}

  \twocolumn[
  \icmltitle{Hybrid Associative Memories}

  \begin{icmlauthorlist}

 \icmlauthor{Leon Lufkin}{zyphra,core}
 \icmlauthor{Tomas Figliolia}{zyphra}
 \icmlauthor{Beren Millidge}{zyphra}
 \icmlauthor{Kamesh Krishnamurthy}{zyphra,core,lead}
  \end{icmlauthorlist}

  \icmlaffiliation{zyphra}{Zyphra}

  \icmlcorrespondingauthor{Kamesh Krishnamurthy}{kamesh@zyphra.com}
  \icmlcorrespondingauthor{Leon Lufkin}{leon@zyphra.com}

  \vskip 0.1in
  {\centering\small\textsuperscript{1}Zyphra\par}
  \vskip 0.0375in
  {\centering\small\textsuperscript{$\dagger$}Project lead\par}
  \vskip 0.0375in
  {\centering\small\textsuperscript{*}Corresponding authors: \texttt{\{kamesh,leon\}@zyphra.com}\par}
  \vskip 0.3in
  ]


  \begin{abstract}

  Recurrent neural networks (RNNs) and self-attention are both widely used sequence-mixing layers that maintain
   an internal memory. However, this memory is constructed using two orthogonal mechanisms: RNNs compress the entire past into a fixed-size state,
  whereas self-attention's state stores every past time step growing its state (the KV cache) linearly with the sequence length. This results in orthogonal strengths and weaknesses. Self-attention layers excel at retrieving information in the context but have large memory and computational costs, while RNNs
  are more efficient but degrade over longer contexts and underperform for precise recall tasks.
  Prior work combining these mechanisms has focused primarily on naively interleaving them to reduce computational cost without regard to their complementary mechanisms.
  We propose the Hybrid Associative Memory (HAM) layer, which combines self-attention and RNNs while leveraging their individual strengths: the RNN
  compresses the entire sequence, while attention supplements it \emph{only} with information that is
  difficult for the RNN to predict, which is hence  the most valuable information to explicitly store.
  HAM layers enable data-dependent growth of the KV cache, which can be precisely controlled by the user with a single, continuous threshold. We find that this fine-grained control of the KV cache growth rate has a smooth trade-off with loss and performance.
  Empirically, we show that our hybrid architecture  offers strong, competitive performance relative to RNNs and Transformers even at substantially lower KV-cache usage.

  \end{abstract}

  \section{Introduction}
  The Transformer architecture based on the self-attention mechanism \cite{bahdanau2014neural,vaswani2017attention} forms the basis for most modern LLM architectures. 
Self-attention enables rich sequence modeling via contextual processing of every input by comparing it to all the past inputs. 
This fine-grained processing leads to remarkable performance but comes at a substantial cost: 1) this computation scales quadratically with sequence length, $\mathcal{O}(T^2)$, and 2) the context memory (KV cache) grows linearly with sequence length, $\mathcal{O}(T)$, making long-context inference expensive and eventually impractical. 
These fundamental scaling bottlenecks persist despite efficient methods to implement self-attention \cite{ainslie2023gqa,liu2024deepseek,liu2025deepseek,kwon2023efficient}.

Recently, a family of sequence-processing blocks based on RNNs has started to become competitive with self-attention.
These RNN layers update a fixed-size internal state for every input where this state acts as compressed summary of past inputs.
This allows for computation that is linear in the sequence length ($\mathcal{O}(T)$) and a memory cost that is constant ($\mathcal{O}(1)$).
Due to the requirement of efficient parallel training, most of these modern RNN-based models can be succinctly formulated as gated linear dynamical systems \cite{yang2024gated,yang2024parallelizing,dao2024transformers,peng2024eagle} and include variants that are either approximations to the Softmax kernel or based on state-space models \cite{dao2024transformers,gu2024mamba}, or implement a form of online regression \cite{wang2025test,schlag2021linear,yang2024gated, siems2025deltaproduct,grazzi2024unlocking}\footnote{But also see \cite{behrouz2024titans,behrouz2025s}, for RNN layers doing online, nonlinear regression.}.
These modern RNN models have GPU-friendly, parallelized implementations that allow them to scale efficiently \cite{gu2024mamba,yang2024fla,beck2025tiled}, and they have proven to be more competitive with Transformers relative to older RNN models.
However, in spite of these advances, the lossy, fixed-size RNN state ultimately leads to a fundamental performance gap vs attention in tasks that require precise in-context recall in for long sequences \cite{arora2023zoology,arora2024simple,bick2025understanding}. 

To maintain the high-resolution recall of the Transformers but still leverage the computational efficiency of RNNs, a family of hybrid models has emerged that combine self-attention and RNN in different ways. 
One class of hybrids follows a strategy of interleaving RNN layers with attention layers \cite{glorioso2024zamba,glorioso2024zamba2,Samba2024,team2025kimi}. 
While interleaving reduces the number of attention layers, it does not make more than a constant-factor improvement in the fundamental quadratic bottleneck of the self-attention layers or address the lossy nature of the RNNs. 
More recently, a family of hybrid models implements a ``hybrid head'' where the KV cache and RNN are combined within a single layer \cite{zuo2025falcon,Dong2024Hymba}. 
The parallel processing of every input token by the RNN and the input token at each layer allows the layer to leverage both the former's efficient summarization and the latter's high-resolution recall.
However, this does not mitigate the fundamental computational bottlenecks of quadratic attention and the KV cache.

These prior hybrid approaches do not exploit the orthogonal and potentially complementary mechanisms of the RNN (efficient summarization of the context) and the KV cache (storing every detail in the context).
Thus, in this paper we introduce the Hybrid Associative Memory (HAM) layer which combines the KV cache and the RNN state in a single layer in a complementary fashion: the RNN captures the predictable and compressible content of the context, and the KV cache augments the RNN by {\it only} storing tokens that are hard for the RNN to predict---akin to a notebook.
In the HAM layer, the number of tokens in the KV cache grows as a function of the structure in the input and can be precisely controlled at runtime.
Our framework aligns well with the rich \emph{Complementary Memory Systems} theories in neuroscience in which distinct subsystems handle fast, episodic recall and slower, abstract integration of experience \cite{McClelland1995,OReilly2014,mcclelland2020integration,Sun2023,Kirkpatrick2017}.
As we elaborate below, the HAM architecture offers competitive performance compared to Transformers and modern RNNs in tasks, compute and memory.

Our contributions in this paper are:
\begin{enumerate}
    \item Introduce a novel HAM layer that combines RNNs and attention to work in a complementary fashion, wherein the KV cache only stores tokens that are surprising for the RNN.
    \item We show that HAM allows precise and flexible control of the KV-cache growth rate along the sequence, and thus, enables a continuous interpolation of performance between the full attention and fixed-state RNN
    \item We show that HAM delivers strong, competitive performance relative to transformers even at substantially lower KV-cache usage, outperforms SOTA RNNs, and is comparable to existing approaches to building hybrid models.
    \item We provide a detailed analysis of the internal workings of HAM to reveal the KV-cache growth and how the routing to KV-cache behaves. 
\end{enumerate}

  \section{HAM: A Complementary Hybrid Associative Memory Layer}
  \label{sec:ham-design}

To motivate the design of our HAM layer, we take the perspective of sequence-mixing layers like RNNs and self-attention as forms of associative memory \cite{ramsauer2020hopfield,zhong2025understanding}.
Specifically, these layers maintain a time-dependent state $S_t$ and every input $x_t$ to the layer both updates $S_t$ and generates an output $o_t =\mathcal{F}(S_t,x_t)$ where $\mathcal{F(\cdot)}$ is an associative retrieval function.
In the case of self-attention, we have
\begin{align} \label{eq:kv_output}
    S^{KV}_t & =  \{(k_i,v_i)\}_{i=1}^t \\
    o^{KV}_t & = \mathcal{F}(S^{KV}_t,x_t) = \frac{\sum_{i\le t} \exp(q_t^\top k_i)\, v_i}{\sum_{i\le t} \exp(q_t^\top k_i)}
\end{align}

where $\{q,k,v\}_t$ are the queries, keys and values for the input $x_t$, and $\mathcal{F}$ uses the exponential of the key-query similarity for $\text{softmax}$ retrieval.
As mentioned, this state $S_t$ stores all the past keys and values precisely and grows linearly with the sequence length $T$.

RNN sequence-mixing layers solve the associative recall problem quite differently. 
To illustrate this clearly, let us consider the {\it Linear Attention} (LA) layer which is obtained by converting the exponential kernel of the  $\text{Softmax}$ to a (possibly transformed) dot product \cite{katharopoulos2020transformers}\footnote{Specifically, $\exp(q_t^\top k_i) \approx \phi(q_t)^\top \phi(k_i)$, and for Linear Attention, $\phi(x)=x$.}. 
This leads to a fixed-size matrix-valued memory state which is updated as a (linear) RNN and used as follows:
\begin{align} \label{eq:la_output}
    S^{LA}_t = & S^{LA}_{t-1} + v_t k_t^\top \\
    o^{LA}_t = & S^{LA}_t q_t 
\end{align}
Thus, the RNN state compresses all the past keys and values into a fixed-size state.
This additive compression leads to degraded retrieval due to interference from irrelevant keys.
E.g., if the key relevant to the query $q_t$ is $k_j$ then for long sequences the output will contain the retrieved value and a noise term that can be large:
\begin{align}
    S^{LA}q_t = v_j (q_t^\top k_j) + \overbrace{\sum_{i \neq j} v_i (q_t^\top k_i)}^{\text{noise}}
    \label{eq:interference}
\end{align}

This noise term grows with the sequence length $T$, and is the primary cause of retrieval degradation.
From prior work on Hopfield-like associative memories \cite{amit1987statistical,hopfield1982neural,lucibello2024exponential}, we know that for a  state with a key dimension $d_k$ the LA retrieval will start to rapidly degrade beyond a sequence length $T=O(d_k)$.
The recall from the KV cache in self-attention is vastly improved since the $\text{softmax}$ exponentially sharpens similarities and thus significantly reduces interference noise, at the cost of a growing state size \citep{krotov2016dense,millidge2022universal}.

\paragraph{Reducing Interference via Online Regression: DeltaNet} 
The DeltaNet architecture \cite{schlag2021linear,yang2024parallelizing} partially addresses the memory interference issue by framing the RNN state update as an online regression problem. 
Specifically, the DeltaNet state $S_t$ update seeks to lower a loss $\mathcal{L}(S_t)$ which measures how similar a new key-value pair is to previous pairs stored in the state, and if the similarity is large then the state is not updated much.
This intuitively makes sense since otherwise adding the same pairs to the fixed-size state would make it grow without bound.
Specifically, the DeltaNet update and objective are given by:
\begin{align}
    \mathcal{L}_{t}(S) & =  \tfrac{1}{2} \| S k_{t} - v_{t} \|^2 \\
    S_t & = S_{t-1} - \beta_t \nabla_{S} \mathcal{L}_t(S_{t-1}) \\
  \implies  S_t & = S_{t-1}\bigl(I-\beta_t k_t k_t^\top\bigr) + \beta_t v_t k_t^\top
    \label{eq:deltanet_update}
\end{align}

where $\beta_t$ is the stepwise ``learning rate''.
In spite of DeltaNet's (and its gated variant's) superior performance  compared to Linear Attention, these linear RNN models cannot overcome the fundamental bottleneck where retrieval degrades when $T$ becomes larger than $\mathcal{O}(d_k)$.

\paragraph{Combining the KV Cache and RNN}
The RNN and KV cache offer complementary inductive biases: RNN-style summarization (including DeltaNet-like updates) can capture regularities over long contexts which might aid generalization,
but it inevitably trades off \emph{precision} for \emph{compression}.
Conversely, a KV cache retains every detail but does not enforce a compact representation of the context.
This means that the KV cache representations may generalize less efficiently and also can introduce noise from specific tokens which get averaged out in the RNN state.
Empirically, we observe RNN-based models outperform on short-context sequence modelling metrics and loss compared to self-attention but fall behind in deeper reasoning and precise retrieval over long contexts.
Existing hybrid models either stack these layers or combine them in parallel but do not exploit their complementary strengths; rather, both components process all tokens independently.
Instead, we propose instead a \textit{complementary} hybrid where the RNN captures regularities and the KV cache serves as a \textit{scratchpad} that stores only tokens difficult for the RNN to predict and thus potentially likely to increase interference for the RNN, while removing a large number of unnecesssary or predictable tokens from the expensive KV-cache state.

\paragraph{Prediction Error as Routing Metric}
The choice of which tokens to route to the KV cache is defined by a routing metric.
Given the success of the DeltaNet variants, as a starting point, we use the prediction error used for the DeltaNet update as the routing metric.
Specifically, we compute the prediction error $e_t^{(h)}$ at each time-step as the discrepancy between the RNN's prediction and the target value.
For each head $h$ at time $t$:
\begin{equation}
    e_t^{(h)} = \mathcal{D}\left(S_{t-1}^{(h)} k_t^{(h)}, \, v_t^{(h)}\right)
    \label{eq:fit_error}
\end{equation}
where $\mathcal{D}$ is a distance metric (e.g., cosine distance\footnote{We base the use of the cosine distance instead of a normalized MSE based on the typical values of $\|S_{t-1} k_t \|$ and $\|v_t\|$.}). 
We route tokens to the KV cache using a threshold $\tau$ that can be learned end-to-end or adjusted to control the KV cache budget.
A token is selected if all heads' prediction errors exceed the threshold:
\begin{equation}
\label{eq:routing}
e_t = \min_h e_t^{(h)}, \quad
m_t = \mathbb{I}\bigl[ e_t \ge \tau\bigr].
\end{equation}
The scratchpad stores only selected pairs
$\{(k_i,v_i): m_i=1\}$ and answers queries using standard attention restricted to this subset:
\begin{equation}
\label{eq:scratchpad_attn}
o_t^{\mathrm{KV}}
=
\mathrm{Attn}\Bigl(q_t,\ \{k_i\}_{m_i=1},\ \{v_i\}_{m_i=1}\Bigr).
\end{equation}

Let $o_t^{\mathrm{RNN}}$ denote the RNN readout and $o_t^{\mathrm{KV}}$ the KV cache readout.
We normalize each head output (RMSNorm) and combine via input-dependent head-wise gates:
\begin{align}
\label{eq:ham_gate}
\tilde o_t^{\mathrm{RNN}} =\mathrm{Norm}_h(o_t^{\mathrm{RNN}}),&\:
\tilde o_t^{\mathrm{KV}}=\mathrm{Norm}_h(o_t^{\mathrm{KV}}) \\
o_t = g^{RNN}_t\cdot \tilde o_t^{\mathrm{RNN}} +& g_t^{KV}\cdot \tilde o_t^{\mathrm{KV}},
\end{align}
This design encourages the RNN path to explain as much as possible, while the scratchpad contributes selectively when useful.
The KV cache thus grows in a data-dependent manner and can be controlled via the threshold $\tau$, and the computational cost grows with the number of surprising tokens in the KV cache (instead of $T$). We define $\rho_{KV}=T_{KV}/T$ to denote the KV cache usage for a sequence.

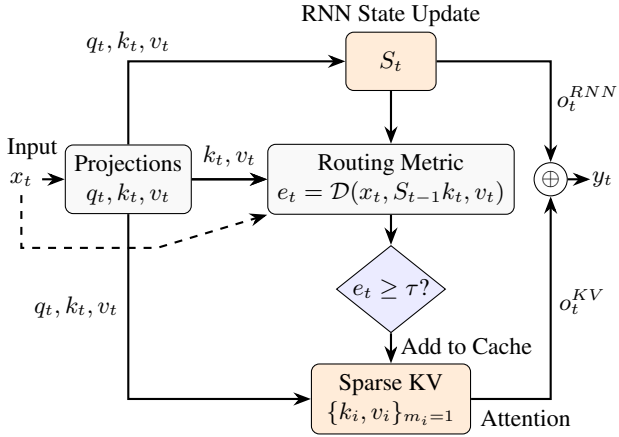
\begin{figure}[t]
\centering
\resizebox{\columnwidth}{!}{%
\begin{tikzpicture}[
    font=\small,
    node distance=7mm and 10mm,
    line/.style={-Stealth, thick},
    label/.style={font=\small, fill=white, fill opacity=0.95, text opacity=1, inner sep=1.1pt, outer sep=0.6pt},
    block/.style={draw, rectangle, rounded corners=3pt, minimum height=2.1em, minimum width=3.4em, align=center, fill=gray!5, inner sep=3.5pt},
    state/.style={draw, rectangle, rounded corners=3pt, minimum height=2.2em, minimum width=3.4em, align=center, fill=orange!15, inner sep=3.5pt},
    decision/.style={draw, diamond, aspect=1.25, fill=blue!8, align=center, inner sep=0.6pt},
    sum/.style={draw, circle, inner sep=0pt, minimum size=1.25em}
]

\node (xt) {$x_t$};
\node[block, right=3mm of xt] (projections) {Projections\\[1pt]{$q_t, k_t, v_t$}};
\node[block, right=of projections, minimum width=8.5em] (fit_error) {Routing Metric\\[1pt]{$e_t = \mathcal{D}(x_t, S_{t-1}k_t, v_t)$}};

\node[state, above=of fit_error] (rnn_state) {$S_t$};
\node[anchor=south, align=center, font=\small] at ([yshift=-0.5mm]rnn_state.north) {RNN State Update};

\node[decision, below=4mm of fit_error] (decide) {$e_t \geq \tau$?};
\node[state, below=4mm of decide, minimum width=6.0em] (kv_cache) {Sparse KV\\[1pt]{$\{k_i, v_i\}_{m_i=1}$}};
\node[anchor=west, align=left, ] at ([xshift=0mm, yshift=-2.5mm]kv_cache.east) {Attention};

\node[sum, right=2.5mm of fit_error] (merge) {$\oplus$};
\node[right=2mm of merge] (output) {$y_t$};

\draw[line] (xt) -- node[label, midway, above, yshift=6pt, xshift=-8pt] {Input} (projections);
\draw[line] (projections) -- (fit_error);

\draw[line] (projections.north) |- node[label, pos=0.35, above, yshift=11pt, fill opacity=0] {$q_t, k_t, v_t$} (rnn_state.west);
\draw[line] (rnn_state) -- (fit_error);

\draw[line] (projections) -- node[label, midway, above, yshift=3pt] {$k_t, v_t$} (fit_error);

\draw[line] (fit_error) -- (decide);
\draw[line] (decide) -- node[label, midway, right, xshift=2pt, yshift=1pt] {Add to Cache} (kv_cache);

\draw[line] (projections.south) |- node[label, pos=0.25, left, xshift=-2pt] {$q_t, k_t, v_t$} (kv_cache.west);

\draw[line] (rnn_state.east) -| node[label, near end, above, yshift=-1.5pt, xshift=14pt, fill opacity=0] {$o_t^{RNN}$} (merge.north);
\draw[line] (kv_cache.east) -| node[label, near end, below, yshift=5pt, xshift=12pt, fill opacity=0] {$o_t^{KV}$} (merge.south);

\draw[line] (merge) -- ++(5.25mm,0);

\draw[line, dashed] (xt.south) -- ++(0,-7mm) -- ++(18mm,0) -- (fit_error.south west);

\end{tikzpicture}%
}
\caption{Hybrid Associative Memory that maintains both an RNN state $S_t$ and KV cache which stores surprising tokens. The two memories work together in a complementary fashion.}
\label{fig:ham-schematic}
\end{figure}

\paragraph{Learned routing metrics}
Some of the more complex RNN models can be viewed as linear attention with a learned operator \cite{liu2026test}.
This motivates us to consider a learned routing operator $\mathcal{D}_{\theta}$, in addition to the prediction error.
In this case, we define the routing score as $  e_t^{(h)} = \mathcal{D}_{\theta}\left(S_{t-1}^{(h)} k_t^{(h)}, \, v_t^{(h)},x_t\right)$.
We parametrize this learned routing operator as a MLP with a sigmoid at the end.
We further simplify this learned router by only feeding it the input $x_t$, and thus the error is given by
\begin{equation}
\label{eq:learned-router}
e_t = \sigma\left( \textrm{MLP}(x_t) \right)
\end{equation}

\paragraph{Implementation details}
The RNN and KV cache share the $K,Q,V$ projection weights, but all the gating-related parameters are independent\footnote{We found that having separate projections offers no real benefit for a substantial increase in the number of parameters.}.
The RNN is implemented using flash-linear-attention \cite{yang2024fla} kernels that are modified to handle the routing.
The attention over the (dynamic) sparse KV cache is implemented using FlexAttention with a custom BlockMask constructor.
More details about the design and implementation of the HAM model are give in Appendix \ref{sec:arch_details}.

  \section{Related Work}
  
\paragraph{Stacked Hybrid Architectures.}
Early attempts to overcome the quadratic scaling of Transformers focused on interleaving or stacking different sequence-mixing layers. 
These layer-wise hybrids typically alternate between a linear-time module, such as a State-Space Model (SSM) or a Recurrent Neural Network (RNN), and a standard self-attention layer \cite{glorioso2024zamba, ren2024samba}. 
For instance, Jamba \cite{team2025jamba} and Samba \cite{ren2024samba} utilize this strategy to achieve long-context efficiency. 
Some recent hybrid \cite{glorioso2024zamba2,ibm2025granite} employ a serial hybrid design where the vast majority of layers are Mamba-2 blocks, with only occasional attention layers inserted to provide ``precision fixes.'' 
While these models reduce the overall computational footprint, they suffer from a fundamental trade-off: the RNN layers remain lossy, and the attention layers still require a full KV cache for every token they process \cite{hyMBA2024}. 
Consequently, the model may encounter bottlenecks if one layer type cannot adequately compensate for the weaknesses of another.

\paragraph{Hybrid-Head Models.}
To achieve a more synergistic integration, a family of ``hybrid-head'' architectures has emerged, combining attention and recurrence within a single layer. 
In these models, such as NVIDIA's HyMBA \cite{hyMBA2024} and TII's Falcon-H1 \cite{falconh1_2025}, self-attention heads and SSM/RNN units (like Mamba or Gated DeltaNet) run in parallel on the same input activations. 
This intra-layer fusion allows the model to leverage high-resolution recall and compressed long-term memory simultaneously. 
HyMBA combines the RNN output with the KV cache output additively, whereas Falcon-H1 concatenates the outputs. 
Other notable efforts include the Dragon LLM \cite{dragon2025} and Nemotron-9B \cite{nemotron2025}, which integrate Gated Delta Networks (GDN). 
However, even in these parallel designs, the KV cache and the RNN typically function independently; both modules process every token, and the KV cache continues to grow linearly with the total sequence length. 
In \cite{irie2025blending}, the authors use a fixed-size KV cache to augment an RNN, but the tokens routed to the KV cache are done in fixed manner without any relation to the RNN state.

A recent alternative approach is the DeepSeek DSA layer which uses a cheaper single-head attention to select the top-k tokens for the more expensive full attention to attend to \cite{deepseekai2025deepseekv32exp}.
This approach reduces the cost of  the full attention to be RNN-like since the `state' is dynamically constructed per token based on the top-k tokens selected by the indexer.
However,  the entire KV cache needs to be stored at each layer and the indexer itself is quadratic in compute, so this approach does not address the fundamental  memory and compute bottlenecks of attention. Another approach which uses a learnable router to select tokens in a sequence is H-net \citep{hwang2025dynamic}.
This approach is used primarily to make byte tokenization more feasible.  Here multiple SSMs act as a router to decide which tokens to select, which are then processed by the main model. However, unlike HAM this selection is only done at the beginning of the model, whereas in HAM the selection is  applied at every layer, enabling dynamic compressions per layer, and unlike HAM, HNet has no complementary storage mechanism for tokens that are dropped


  \section{Experiments}
  \subsection{Model Comparison}
\label{subsec:model-comparison}


\begin{table*}[t]
\caption{Model comparison at 800M scale. \emph{All HAM models use 50\% KV cache ($\rho_{KV}=0.5$).}}
\label{tab:model_comparison}
\vskip -0.02in

\begin{subtable}{\textwidth}
\subcaption{Standard language-modeling and zero-shot commonsense reasoning benchmarks.}
\label{tab:model_comparison_standard}
\begin{center}
\begin{small}
\begin{tabular}{lccccccc|cc|c}
\toprule
Model     & Hella. & PIQA & ARC-e & ARC-c & Wino. & BoolQ & LMB. & Wiki. & LMB. & Avg \\
 & acc\_n $\uparrow$ & acc $\uparrow$ & acc\_n $\uparrow$ & acc\_n $\uparrow$ & acc $\uparrow$ & acc $\uparrow$ & acc $\uparrow$ & ppl $\downarrow$ & ppl $\downarrow$ & $\uparrow$ \\
\midrule
Transformer     & 43.3 & 67.4 & 48.4 & \underline{27.7} & 51.5 & \underline{58.3} & \textbf{44.9} & \underline{19.66} & \textbf{13.44} & 48.8 \\
GDN     & 43.2 & 66.2 & 47.3 & 27.2 & 52.2 & 51.7 & 42.4 & 22.91 & 15.41 & 47.2 \\
\midrule
GDN-GSA     & \textbf{44.7} & \textbf{68.2} & 48.9 & \textbf{29.0} & \textbf{53.7} & 50.2 & 44.0 & 22.04 & 14.39 & 48.4 \\
\midrule
\emph{HAM} \\
Fixed $\tau$     & 43.8 & 66.2 & 47.9 & 26.5 & 52.8 & 57.7 & 43.7 & 19.85 & 14.16 & 48.4 \\
Learned $\tau$     & 43.8 & 65.6 & 49.7 & 27.0 & \underline{53.6} & 56.2 & 42.7 & 19.94 & 15.67 & 48.4 \\
Learned router     & 43.8 & \underline{67.9} & \textbf{50.8} & 27.3 & 52.6 & 56.5 & \underline{44.6} & 20.15 & \underline{13.96} & \textbf{49.1} \\
Learned router + EDA     & \underline{44.1} & 67.1 & \underline{50.6} & 26.9 & 52.8 & \textbf{59.7} & 42.8 & \textbf{19.49} & 14.90 & \textbf{49.1} \\
\bottomrule
\end{tabular}
\end{small}
\end{center}
\end{subtable}

\vskip 0.075in

\begin{subtable}{\textwidth}
\subcaption{Long-context evaluation results on RULER tasks.}
\label{tab:model_comparison_longctx}
\centering
\providecommand{\sz}[1]{{\small #1}}
\resizebox{\textwidth}{!}{%
\setlength{\tabcolsep}{1.5pt}
\begin{tabular}{lc@{\hskip 2pt}c@{\hskip 2pt}c|c@{\hskip 2pt}c@{\hskip 2pt}c|c@{\hskip 2pt}c@{\hskip 2pt}c|c@{\hskip 2pt}c@{\hskip 2pt}c|c@{\hskip 2pt}c@{\hskip 2pt}c|c@{\hskip 2pt}c@{\hskip 2pt}c|c@{\hskip 2pt}c@{\hskip 2pt}c|c@{\hskip 2pt}c@{\hskip 2pt}c|c@{\hskip 2pt}c@{\hskip 2pt}c|c@{\hskip 2pt}c@{\hskip 2pt}c|c@{\hskip 2pt}c@{\hskip 2pt}c}
\toprule
 & \multicolumn{3}{c|}{S1} & \multicolumn{3}{c|}{S2} & \multicolumn{3}{c|}{S3} & \multicolumn{3}{c|}{MK1} & \multicolumn{3}{c|}{MK2} & \multicolumn{3}{c|}{MQ} & \multicolumn{3}{c|}{MV} & \multicolumn{3}{c|}{CWE} & \multicolumn{3}{c|}{FWE} & \multicolumn{3}{c|}{QA-H} & \multicolumn{3}{c}{QA-S} \\
\cmidrule(lr){2-4}\cmidrule(lr){5-7}\cmidrule(lr){8-10}\cmidrule(lr){11-13}\cmidrule(lr){14-16}\cmidrule(lr){17-19}\cmidrule(lr){20-22}\cmidrule(lr){23-25}\cmidrule(lr){26-28}\cmidrule(lr){29-31}\cmidrule(lr){32-34}
Model & 4k & 8k & 16k & 4k & 8k & 16k & 4k & 8k & 16k & 4k & 8k & 16k & 4k & 8k & 16k  & 4k & 8k & 16k & 4k & 8k & 16k  & 4k & 8k & 16k & 4k & 8k & 16k & 4k & 8k & 16k & 4k & 8k & 16k \\
\midrule
Transformer
 & \sz{100} & \sz{100} & \sz{98} & \sz{\textbf{100}} & \sz{\textbf{100}} & \sz{99} & \sz{\underline{90}} & \sz{\textbf{95}} & \sz{\textbf{92}} & \sz{\underline{90}} & \sz{\underline{87}} & \sz{61} & \sz{\underline{28}} & \sz{\underline{37}} & \sz{\underline{14}} & \sz{49} & \sz{41} & \sz{35} & \sz{44} & \sz{44} & \sz{28} & \sz{6.8} & \sz{0.6} & \sz{0.2} & \sz{\textbf{36}} & \sz{\textbf{37}} & \sz{\textbf{24}} & \sz{27} & \sz{\underline{28}} & \sz{21} & \sz{\textbf{42}} & \sz{\underline{27}} & \sz{\underline{19}} \\
GDN
 & \sz{100} & \sz{100} & \sz{100} & \sz{\textbf{100}} & \sz{66} & \sz{21} & \sz{52} & \sz{24} & \sz{3.0} & \sz{37} & \sz{31} & \sz{17} & \sz{2.0} & \sz{0.4} & \sz{0.2} & \sz{27} & \sz{24} & \sz{16} & \sz{22} & \sz{22} & \sz{13} & \sz{23} & \sz{6.7} & \sz{0.6} & \sz{13} & \sz{11} & \sz{17} & \sz{22} & \sz{20} & \sz{17} & \sz{19} & \sz{12} & \sz{8.8} \\
\midrule
GDN-GSA
 & \sz{100} & \sz{100} & \sz{100} & \sz{\textbf{100}} & \sz{\textbf{100}} & \sz{\textbf{100}} & \sz{\textbf{93}} & \sz{\underline{86}} & \sz{\underline{48}} & \sz{83} & \sz{86} & \sz{\textbf{77}} & \sz{15} & \sz{12} & \sz{2.6} & \sz{\textbf{72}} & \sz{\underline{62}} & \sz{\textbf{39}} & \sz{\underline{59}} & \sz{\underline{57}} & \sz{\textbf{33}} & \sz{12} & \sz{0.2} & \sz{0.0} & \sz{28} & \sz{12} & \sz{0.9} & \sz{\textbf{37}} & \sz{\textbf{31}} & \sz{\textbf{27}} & \sz{36} & \sz{23} & \sz{\underline{19}} \\
\midrule
\emph{HAM} \\
Fixed $\tau$
 & \sz{100} & \sz{100} & \sz{100} & \sz{99} & \sz{94} & \sz{57} & \sz{2.0} & \sz{12} & \sz{1.8} & \sz{39} & \sz{28} & \sz{15} & \sz{0.8} & \sz{0.4} & \sz{0.0} & \sz{35} & \sz{22} & \sz{10} & \sz{20} & \sz{14} & \sz{6.0} & \sz{\underline{56}} & \sz{\textbf{37}} & \sz{\underline{1.1}} & \sz{28} & \sz{28} & \sz{11} & \sz{27} & \sz{22} & \sz{11} & \sz{26} & \sz{16} & \sz{12} \\
Learned $\tau$
 & \sz{100} & \sz{100} & \sz{99} & \sz{95} & \sz{90} & \sz{68} & \sz{27} & \sz{4.6} & \sz{7.8} & \sz{42} & \sz{38} & \sz{31} & \sz{0.6} & \sz{0.2} & \sz{0.0} & \sz{24} & \sz{22} & \sz{16} & \sz{28} & \sz{26} & \sz{21} & \sz{4.3} & \sz{0.5} & \sz{0.3} & \sz{\underline{32}} & \sz{\underline{32}} & \sz{\underline{18}} & \sz{26} & \sz{25} & \sz{21} & \sz{22} & \sz{16} & \sz{12} \\
Learned router
 & \sz{100} & \sz{100} & \sz{100} & \sz{78} & \sz{\textbf{100}} & \sz{72} & \sz{87} & \sz{60} & \sz{22} & \sz{\textbf{95}} & \sz{\textbf{90}} & \sz{69} & \sz{\textbf{67}} & \sz{\textbf{45}} & \sz{\textbf{19}} & \sz{\underline{70}} & \sz{\textbf{64}} & \sz{\underline{36}} & \sz{\textbf{82}} & \sz{\textbf{74}} & \sz{\underline{32}} & \sz{\textbf{58}} & \sz{\underline{26}} & \sz{\textbf{2.1}} & \sz{1.0} & \sz{1.1} & \sz{8.9} & \sz{\underline{31}} & \sz{26} & \sz{\underline{23}} & \sz{\textbf{42}} & \sz{\textbf{28}} & \sz{\textbf{23}} \\
Learned router + EDA
 & \sz{100} & \sz{100} & \sz{100} & \sz{78} & \sz{\textbf{100}} & \sz{\textbf{100}} & \sz{43} & \sz{37} & \sz{20} & \sz{59} & \sz{79} & \sz{\underline{75}} & \sz{27} & \sz{18} & \sz{9.2} & \sz{36} & \sz{33} & \sz{28} & \sz{30} & \sz{29} & \sz{26} & \sz{31} & \sz{0.9} & \sz{0.0} & \sz{3.4} & \sz{2.6} & \sz{2.7} & \sz{28} & \sz{22} & \sz{22} & \sz{36} & \sz{26} & \sz{\underline{19}} \\
\bottomrule
\end{tabular}%
}
\end{subtable}

\vskip -0.1in
\end{table*}


\begin{table*}[t]
\caption{$T_\mathrm{KV}/T$ sweep results at 800M scale. $^*$GDN baseline (no KV).}
\label{tab:ham_sweep}
\vskip -0.02in

\begin{subtable}{\textwidth}
\subcaption{Standard language-modeling and zero-shot commonsense reasoning benchmarks.}
\label{tab:ham_sweep_standard}
\begin{center}
\begin{small}
\begin{tabular}{lccccccc|cc|c}
\toprule
$\sfrac{T_\mathrm{KV}}{T}$    & Hella. & PIQA & ARC-e & ARC-c & Wino. & BoolQ & LMB. & Wiki. & LMB. & Avg \\
 & acc\_n $\uparrow$ & acc $\uparrow$ & acc\_n $\uparrow$ & acc\_n $\uparrow$ & acc $\uparrow$ & acc $\uparrow$ & acc $\uparrow$ & ppl $\downarrow$ & ppl $\downarrow$ & $\uparrow$ \\
\midrule
$0^*$    & 43.2 & 66.2 & 47.3 & 27.2 & 52.2 & 51.7 & 42.4 & 22.91 & 15.41 & 47.2 \\
\midrule
$\sfrac{1}{4}$    & 43.2 & 67.1 & 46.8 & \underline{28.6} & 52.7 & \underline{58.5} & 41.7 & 20.21 & 16.03 & \textbf{48.4} \\
$\sfrac{3}{8}$    & 43.3 & \textbf{68.3} & 47.9 & 27.4 & \textbf{54.7} & 50.1 & 41.0 & 20.03 & 15.87 & 47.5 \\
$\sfrac{1}{2}$    & 43.8 & 65.6 & \textbf{49.7} & 27.0 & \underline{53.6} & 56.2 & 42.7 & 19.94 & 15.67 & \textbf{48.4} \\
$\sfrac{5}{8}$    & 43.5 & 65.6 & 47.6 & 27.8 & 53.4 & \textbf{59.7} & 41.5 & 19.95 & 16.33 & \textbf{48.4} \\
$\sfrac{3}{4}$    & \underline{44.2} & \underline{67.4} & \underline{48.0} & \textbf{28.8} & 53.3 & 51.5 & \textbf{44.0} & \underline{19.79} & \underline{14.41} & 48.2 \\
$1$    & \textbf{45.1} & 65.1 & \underline{48.0} & 24.9 & 53.2 & 54.3 & \underline{43.8} & \textbf{19.21} & \textbf{14.22} & 47.8 \\
\bottomrule
\end{tabular}
\end{small}
\end{center}
\end{subtable}

\vskip 0.075in

\begin{subtable}{\textwidth}
\subcaption{Long-context evaluation results on RULER tasks.}
\label{tab:ham_sweep_longctx}
\centering
\providecommand{\sz}[1]{{\small #1}}
\resizebox{\textwidth}{!}{%
\setlength{\tabcolsep}{1.5pt}
\begin{tabular}{lc@{\hskip 2pt}c@{\hskip 2pt}c|c@{\hskip 2pt}c@{\hskip 2pt}c|c@{\hskip 2pt}c@{\hskip 2pt}c|c@{\hskip 2pt}c@{\hskip 2pt}c|c@{\hskip 2pt}c@{\hskip 2pt}c|c@{\hskip 2pt}c@{\hskip 2pt}c|c@{\hskip 2pt}c@{\hskip 2pt}c|c@{\hskip 2pt}c@{\hskip 2pt}c|c@{\hskip 2pt}c@{\hskip 2pt}c|c@{\hskip 2pt}c@{\hskip 2pt}c|c@{\hskip 2pt}c@{\hskip 2pt}c}
\toprule
 & \multicolumn{3}{c|}{S1} & \multicolumn{3}{c|}{S2} & \multicolumn{3}{c|}{S3} & \multicolumn{3}{c|}{MK1} & \multicolumn{3}{c|}{MK2} & \multicolumn{3}{c|}{MQ} & \multicolumn{3}{c|}{MV} & \multicolumn{3}{c|}{CWE} & \multicolumn{3}{c|}{FWE} & \multicolumn{3}{c|}{QA-H} & \multicolumn{3}{c}{QA-S} \\
\cmidrule(lr){2-4}\cmidrule(lr){5-7}\cmidrule(lr){8-10}\cmidrule(lr){11-13}\cmidrule(lr){14-16}\cmidrule(lr){17-19}\cmidrule(lr){20-22}\cmidrule(lr){23-25}\cmidrule(lr){26-28}\cmidrule(lr){29-31}\cmidrule(lr){32-34}
$\sfrac{T_\mathrm{KV}}{T}$ & 4k & 8k & 16k & 4k & 8k & 16k & 4k & 8k & 16k & 4k & 8k & 16k & 4k & 8k & 16k  & 4k & 8k & 16k & 4k & 8k & 16k  & 4k & 8k & 16k & 4k & 8k & 16k & 4k & 8k & 16k & 4k & 8k & 16k \\
\midrule
$0^*$
 & \sz{100} & \sz{100} & \sz{100} & \sz{\textbf{100}} & \sz{66} & \sz{21} & \sz{52} & \sz{24} & \sz{3.0} & \sz{37} & \sz{31} & \sz{17} & \sz{2.0} & \sz{0.4} & \sz{0.2} & \sz{27} & \sz{24} & \sz{16} & \sz{22} & \sz{22} & \sz{13} & \sz{\underline{23}} & \sz{\underline{6.7}} & \sz{\underline{0.6}} & \sz{13} & \sz{11} & \sz{17} & \sz{22} & \sz{20} & \sz{17} & \sz{19} & \sz{12} & \sz{8.8} \\
\midrule
$\sfrac{1}{4}$
 & \sz{100} & \sz{100} & \sz{100} & \sz{95} & \sz{62} & \sz{40} & \sz{64} & \sz{36} & \sz{9.2} & \sz{34} & \sz{26} & \sz{13} & \sz{0.4} & \sz{0.2} & \sz{0.2} & \sz{22} & \sz{15} & \sz{8.0} & \sz{22} & \sz{18} & \sz{7.2} & \sz{6.4} & \sz{0.6} & \sz{0.3} & \sz{9.4} & \sz{4.1} & \sz{4.1} & \sz{25} & \sz{21} & \sz{17} & \sz{19} & \sz{13} & \sz{11} \\
$\sfrac{3}{8}$
 & \sz{100} & \sz{100} & \sz{100} & \sz{\textbf{100}} & \sz{\underline{99}} & \sz{24} & \sz{80} & \sz{\textbf{54}} & \sz{\underline{15}} & \sz{36} & \sz{33} & \sz{10} & \sz{2.4} & \sz{0.4} & \sz{0.0} & \sz{23} & \sz{19} & \sz{12} & \sz{22} & \sz{19} & \sz{12} & \sz{21} & \sz{2.1} & \sz{0.2} & \sz{14} & \sz{17} & \sz{\underline{19}} & \sz{26} & \sz{24} & \sz{17} & \sz{22} & \sz{13} & \sz{10} \\
$\sfrac{1}{2}$
 & \sz{100} & \sz{100} & \sz{99} & \sz{95} & \sz{90} & \sz{68} & \sz{27} & \sz{4.6} & \sz{7.8} & \sz{42} & \sz{38} & \sz{31} & \sz{0.6} & \sz{0.2} & \sz{0.0} & \sz{24} & \sz{22} & \sz{16} & \sz{28} & \sz{26} & \sz{21} & \sz{4.3} & \sz{0.5} & \sz{0.3} & \sz{\textbf{32}} & \sz{\underline{32}} & \sz{18} & \sz{26} & \sz{25} & \sz{21} & \sz{22} & \sz{\underline{16}} & \sz{12} \\
$\sfrac{5}{8}$
 & \sz{100} & \sz{100} & \sz{99} & \sz{\textbf{100}} & \sz{98} & \sz{89} & \sz{\underline{86}} & \sz{37} & \sz{\textbf{36}} & \sz{67} & \sz{57} & \sz{56} & \sz{1.6} & \sz{1.0} & \sz{0.0} & \sz{41} & \sz{39} & \sz{\textbf{31}} & \sz{37} & \sz{30} & \sz{\underline{28}} & \sz{22} & \sz{1.7} & \sz{\textbf{0.7}} & \sz{\underline{30}} & \sz{\textbf{34}} & \sz{\textbf{31}} & \sz{27} & \sz{24} & \sz{16} & \sz{\underline{28}} & \sz{\textbf{17}} & \sz{\underline{14}} \\
$\sfrac{3}{4}$
 & \sz{100} & \sz{100} & \sz{99} & \sz{99} & \sz{98} & \sz{\underline{91}} & \sz{54} & \sz{37} & \sz{7.4} & \sz{\underline{72}} & \sz{\textbf{70}} & \sz{\textbf{61}} & \sz{\underline{3.8}} & \sz{\underline{1.6}} & \sz{\underline{1.4}} & \sz{\underline{59}} & \sz{\underline{54}} & \sz{\underline{30}} & \sz{\underline{42}} & \sz{\underline{39}} & \sz{\textbf{33}} & \sz{\textbf{45}} & \sz{\textbf{11}} & \sz{0.3} & \sz{19} & \sz{9.4} & \sz{8.0} & \sz{\underline{29}} & \sz{\textbf{30}} & \sz{\textbf{22}} & \sz{24} & \sz{\underline{16}} & \sz{\textbf{16}} \\
$1$
 & \sz{100} & \sz{100} & \sz{100} & \sz{\textbf{100}} & \sz{\textbf{100}} & \sz{\textbf{98}} & \sz{\textbf{90}} & \sz{\underline{51}} & \sz{12} & \sz{\textbf{81}} & \sz{\textbf{70}} & \sz{\underline{58}} & \sz{\textbf{41}} & \sz{\textbf{21}} & \sz{\textbf{10}} & \sz{\textbf{76}} & \sz{\textbf{55}} & \sz{18} & \sz{\textbf{76}} & \sz{\textbf{56}} & \sz{18} & \sz{14} & \sz{1.6} & \sz{0.1} & \sz{0.7} & \sz{0.1} & \sz{1.2} & \sz{\textbf{32}} & \sz{\underline{27}} & \sz{\textbf{22}} & \sz{\textbf{29}} & \sz{15} & \sz{13} \\
\bottomrule
\end{tabular}%
}
\end{subtable}

\vskip -0.1in
\end{table*}


\paragraph{Setup}
We first compare HAM against other architectures on standard language modeling and long context benchmarks (\cref{tab:model_comparison}).
We compare the HAM model against the Transformer, GDN, and a hybrid interleaving GDN and {\it full/global} self-attention (GSA) layers, denoted GDN-GSA.
\emph{Note:} this is not the GDN-H1 \cite{yang2024gated}, usually reported, which uses sliding-window attention interleaved with the GDN.
The models are parameter-matched as best as possible with 801M, 804M, 779M, and 805M parameters for the Transformer, GDN, GDN-GSA, and HAM, respectively.
All models have 24 layers and a hidden size of 1,792, except for the Transformer, which has 23 layers and a hidden size of 1,920.
All GDN blocks use key and value head dimensions of 256 and 384, respectively.
Both the Transformer and HAM use a key head dimension 128, but the Transformer uses the same head dimension for values while HAM uses 192 to match the output of the GDN.
We use RoPE with a base of 500K for all inputs to self-attention \cite{su2024roformer}.
All models use SwiGLU MLP blocks \cite{shazeer2020glu} and pre-norm.
Note that the Transformer, GDN, and HAM are \emph{not} FLOPs- or memory-matched; only GDN-GSA and HAM (with half of the KV cache) are matched.
This is inevitable due to the sequence-length dependence of these constraints and the fundamental difference in the architectures.
Precisely, the HAM models (with half of the KV cache), GDN, Transformer, and GDN-GSA hybrid require 0.3511, 0.2467 ($-29.7\%$), 0.4592 ($+30.8\%$), and 0.3429 ($-2.3\%$) zFLOPs to be trained (see \cref{subsubsec:training_flops} for detailed calculations).

All HAM models in \cref{tab:model_comparison} keep 50\% of the KV cache while the GDN-GSA hybrid alternates full attention layers with the same overall KV cache usage as HAM.
We compare four types of HAM models.
The first is a variant with a \emph{fixed} threshold, $\tau$, fixed throughout training that achieves exactly 50\% KV cache usage by the end of training.
The second uses a threshold that is learned and achieves the same KV cache usage by the end of training \cref{sec:arch_details}.
The final two variants use learned routers and also employ learned thresholds to achieve the same final KV cache usage (see \cref{fig:KV_cache_usage}).
The second learned router variant applies exponential decay averaging (EDA) to the routing scores across layers, inspired by the strong performance of EDA in our Zaya-1 model \cite{anthony2025training}, as well as in an attempt to mitigate sudden collapses of deeper layers (c.f \cref{sec:arch_details}).

The models are trained using the Flame training framework \cite{flame2024} on the Long Data Collections dataset \cite{fu2024data} without books \cite{emozilla2024longdata} for 50B tokens.
We use a global batch size of 524K tokens and context length of $16{,}384$ for a total of 95K gradient steps.
Futher training details are provided in \cref{sec:arch_details}
 



\paragraph{Tasks}

We evaluate on standard zero-shot common-sense reasoning benchmarks following prior work~\citep{touvron2023llama,dao2024mamba2}. 
To evaluate models' capabilities at longer context lengths, we evaluate on RULER \cite{hsieh2024ruler}, which tests a model's ability to effectively use its entire context window, up to sequence lengths of 16K tokens.
More details about the task descriptions and setup are provided in \cref{sec:task-details}.
We drop results for NIAH multikey 3 (MK3) and variable tracking (VT) because scores are uniformly low for all models at this scale and this amount of training; the HAM model with a learnable router is the only model to exceed single-digit accuracy.

\paragraph{Standard LM Results: }
On standard zero-shot common-sense benchmarks, HAM models are competitive with and outperform the Transformer baseline, despite retaining only 50\% of the KV cache.
The HAM model with a learned router (without EDA) bests the GDN-GSA hybrid on five of nine standard language-modeling benchmarks and has a slightly higher overall accuracy.
Using a fixed threshold instead of a learned one does not significantly impact performance on any benchmarks.
The HAM models with learnable routers have the highest overall accuracy, even when compared to a Transformer with a full KV cache, indicating more flexible routing mechanisms are key to optimally assigning tokens to the KV cache.

\paragraph{Long-context Results: }
HAM is also competitive at longer contexts despite dropping 50\% of KV tokens.
All models excel at NIAH single 1, since needles are easy to find because the haystack consists of repeated text.
In NIAH single 2, which uses an essay for the haystack, GDN performance falls quickly as sequence length grows, as do the cosine metric-based HAM variants, albeit more slowly.
However, the HAM variants with learnable routers show strong performance, though they show a drop in performance for shorter sequences (4K scores).
We speculate that this is likely due to training on a relatively small number of tokens, and having an end-to-end learnable router for the KV cache is more expressive and hence might require more training to reach its full performance.
All HAM variants struggle on NIAH single 3 relative to the Transformer and GDN-GSA hybrid, which we speculate is due to the fact that the needle is much longer than in single-1 or single-2 (7 vs 32 digits) and may be more challenging to route in its entirety to the KV cache for exact retrieval.
However, the HAM models with learnable routers show strong performance in multikey, multiquery, and multivalue retrieval, always beating the Transformer (often handily), and matching and occasionally substantially exceeding the performance of GDN-GSA, especially on multikey 2.
This suggests HAM has a superior ability to reduce interference during retrieval than full attention, which is precisely what its complementary memory design is supposed to achieve.
However, we recognize there is likely substantial variation in these benchmarks for models at this scale with this amount of training; larger experiments are needed to conclusively determine HAM's benefits.

\paragraph{HAM performance as a function of KV cache usage}
HAM provides flexible control over the size of the KV cache.
To understand how HAM uses its KV cache, we train a suite of identical models using the cosine-based metric that differ only in their target KV cache usage.
In \cref{tab:ham_sweep}, we sweep KV cache usages of size $T_\mathrm{KV} / T = 0, \sfrac{1}{4}, \sfrac{3}{8}, \sfrac{1}{2}, \sfrac{5}{8}, \sfrac{3}{4}$, and $1$, where $T$ denotes the length of the sequence and $T_\mathrm{KV}$ denotes the size of the KV cache.
Here, we treat the GDN as a HAM model with zero KV cache, since its configuration is nearly identical aside from HAM's added parameters for routing to the KV cache.
We evaluate using the same model setup and the same suite of tasks as in \cref{tab:model_comparison}.

We generally observe an increase in scores on standard language modeling benchmarks as KV cache size increases, with the cleanest results on Hellaswag and the perplexity-scored tasks, WikiText and LAMBADA, though trends for other benchmarks are less robust.
For the long context benchmarks, we observe substantial increases in performance across nearly all benchmarks as the KV cache grows.
For instance, NIAH single-2 and NIAH multikey-1 show pronounced improvements at 16k sequence length as KV cache size grows.



  \section{HAM offers a fine-grained control of compute/memory vs. performance trade-off}
\label{sec:ham_tradeoff}

A distinctive feature of the HAM architecture is that it provides a \emph{continuous, fine-grained knob} for controlling the KV cache budget---both per-layer and globally---while gracefully trading off memory usage against language modeling loss and long-context evaluation performance. 
This controllability is a direct consequence of HAM's complementary memory design. 
In contrast, for many prior hybrid sequence models, the memory--performance trade-off is coarse and difficult to control.
Interleaved hybrids trade memory for performance mainly through discrete structural choices---e.g., how many attention layers are present, where they are placed, or how wide the attention window is. 
Hybrid-head models move attention and recurrence into the same layer and does not provide a any control of the KV cache size.

\begin{figure}[ht]
  \centering
  \includegraphics[width=\linewidth]{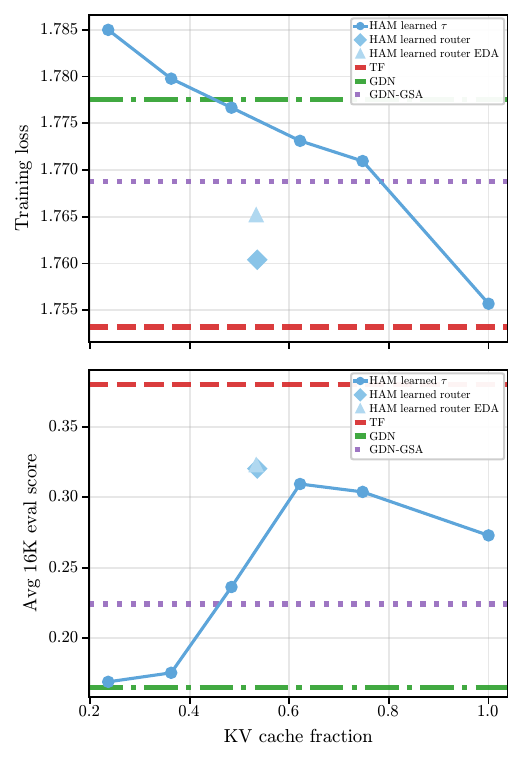}
  \caption{Language modeling loss(top) and long-context accuracy on RULER (bottom) as a function of the KV-cache usage in HAM.  The loss and the accuracy show a smooth relation with the amount of KV-cache HAM uses -- a user-controlled parameter. Notably, the HAM with the learned router with $50\%$ KV-cache usage shows strong performance significantly outperforming the GDN hybrid with full self-attention layers and identical KV-cache usage}
  \label{fig:HAM_loss_evals}
\end{figure}

\paragraph{Per-layer and global control of the KV cache usage.}
HAM makes this control explicit. 
Each layer $\ell$ has its own routing threshold, $\tau_\ell$, which determines which tokens are written to the KV scratchpad and which are handled only by the recurrent state.
This yields two distinct levels of control.
First, \emph{local control}: users can set different thresholds across layers, producing {\it heterogeneous memory profiles} in which some layers retain a larger explicit cache while others rely more heavily on compression. 
Second, \emph{global control}:  the collection $\{\tau_\ell\}_{\ell=1}^L$ can be tuned to match a desired
desired global growth rate $T_{\mathrm{KV}}^* = \rho_{KV} T$, with a single hyperparameter $\rho_{KV}$ that controls the total KV cache size as a fraction of sequence length\footnote{In all our runs, we use a global learnable threshold that achieves a global KV usage target}:
\[
\rho_{\mathrm{KV}} = \frac{1}{LT}\sum_{\ell=1}^L T^{(\ell)}_{\mathrm{KV}}.
\]

Of course, this controllability is only valuable if the resulting trade-off is smooth and predictable.
Empirically, that is exactly what we observe.
In Fig.~\ref{fig:HAM_loss_evals}, we show that this control mechanism yields a smooth trade-off between the KV cache fraction and both training loss and long-context evaluation scores (averaged over NIAH and RULER tasks).
As $T_{\mathrm{KV}}/T$ increases, long-context evaluation improves and language-modeling loss decreases, while lower budgets still retain strong performance because of efficient KV cache compression and the fact that the recurrent pathway continues to summarize the entire past.
This enables practitioners to select an operating point that suits their latency and memory constraints.
Moreover, the learned router, at matched KV fractions, lies above the prediction-error sweep, indicating that HAM can spend a fixed cache budget more efficiently when routing is learned end-to-end.
Thus, HAM offers a precise, layer-wise budget that translates into a usable memory--performance dial.

\paragraph{How HAM differs from prior work on KV cache control}
This level of fine-grained control of the KV-cache size has been difficult to achieve in prior architectures. 
\textit{Stacked hybrids} interleave RNN and attention layers.  
The attention layers in these models maintain a full KV cache over all tokens they process; the only way to reduce KV cache usage is to decrease the \emph{number} of attention layers, which is a discrete, coarse-grained architectural choice made before training.  
There is no mechanism to continuously vary the cache budget within an attention layer. 
\textit{Hybrid-head models} can employ sliding-window attention in certain layers, but these too don't mitigate the KV cache issue. 
\textit{Post-hoc KV cache compression}: A large body of work addresses KV cache reduction as an inference-time optimization applied to pretrained Transformers.  
Methods such as SnapKV~\citep{li2024snapkv}, PyramidKV~\citep{cai2024pyramidkv}, Ada-KV~\citep{feng2024adakv}, LAVa~\citep{lava2025}, and CAKE~\citep{cake2025} provide layer-wise or head-wise budget allocation by evicting tokens deemed unimportant according to attention-score heuristics.
While these methods offer budget control, they differ from HAM in two fundamental ways.
First, they are applied \emph{post hoc} to models that were trained with full KV caches, meaning the model has no opportunity during training to learn to compensate for the missing tokens.  
Second, they lack a complementary compressive memory that can absorb the information carried by evicted tokens; once a token is evicted, its information is lost.
In contrast, HAM's RNN explicitly captures compressible context, and the KV cache is populated only with tokens that the RNN \emph{cannot} predict, ensuring that eviction and compression are coordinated.

HAM has many desirable features in the space of hybrid memories: it preserves the complementary strengths of recurrence and explicit KV recall while converting KV growth into a user-specified resource budget. 
For deployment, this means that the same trained architecture could be potentially steered toward lower-memory or higher-performance regimes without changing the layer type or resorting to external cache-pruning rules.  
More broadly, it provides a clean experimental framework for studying how much explicit memory each layer truly needs, which is much harder to probe in existing hybrid models.
We emphasize that tuning the KV cache using the routing metric and the learnable threshold is extremely flexible. 
For instance, it offers a natural way to impose a ``schedule'' for the threshold where the KV cache growth rate can be varied {\it within a sequence}. 
Moreover, given the relatively smooth trade-off between the KV cache usage and loss/performance, an interesting future direction is to explore varying the threshold during test time.


  \section{Analysis of the internals of HAM}

In this section, we provide more insights into the routing metric and the KV-cache growth in different layers of HAM. 

\subsection{A Toy Example}
\label{subsec:toy}
To illustrate how prediction error can be used to route ``surprising'' tokens to the KV cache, we consider a sequence from the NIAH single 1 benchmark.
The sequence looks like:

\begin{tikzpicture}
\node[draw, rounded corners=5pt, inner sep=8pt, text width=0.95\linewidth, align=center, fill=gray!10] {\small\texttt{The grass is green. The sky is blue. The sun ... }{\color{red}\texttt{One of the special magic numbers for uttermost-sip is: 7441831}}\texttt{ ... The grass is green. The sky is blue. The sun ...}};
\end{tikzpicture}

\begin{figure}[ht]
  \centering
  \includegraphics[width=\linewidth]{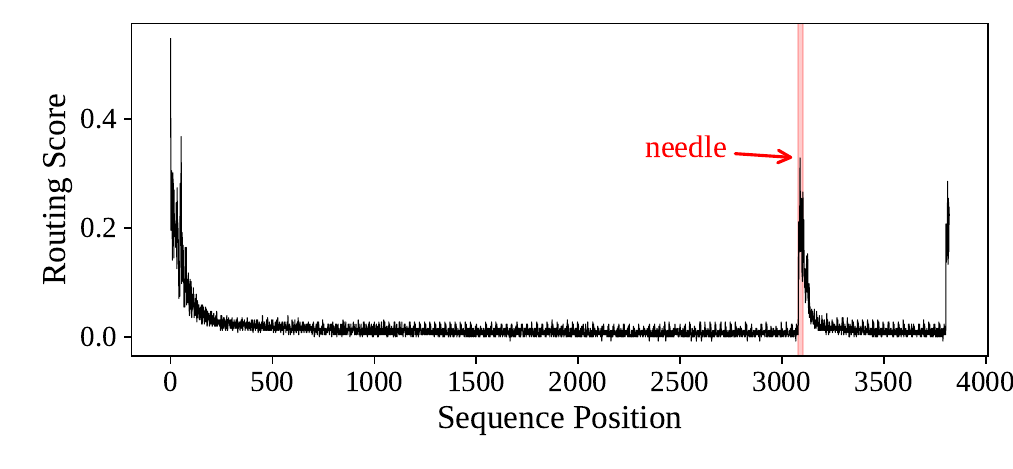}
  \caption{Trajectory of the prediction error for the toy NIAH single 1 sequence. 
  Obtained from the 10th layer of an 800M HAM model using a cosine-similiarity routing score trained on 50B tokens from Long Data Collections.}
  \label{fig:toy}
\end{figure}

In \cref{fig:toy}, we track the routing score for each time $t$ for a given layer, which clearly shows a spike up to nearly 0.4 when the pattern-breaking needle is encountered.
The spike is wide enough to capture the entire needle sentence (23 tokens total).
This increase identifies these tokens for selection into the KV cache.
There is also a smaller spike at the end of the sequence, which corresponds to the prompt to retrieve the needle.
In \cref{fig:toy_learnable}, we see that the end-to-end learnable router shows similar behavior.
More realistic language datasets do not show such clear trends, but we use this example to demonstrate how the routing mechanism identifies tokens which are surprising to the RNN.

\subsection{Routing Score Trends and its Relation to Gating}
\begin{figure}[ht]
  \centering
  \includegraphics[width=\linewidth]{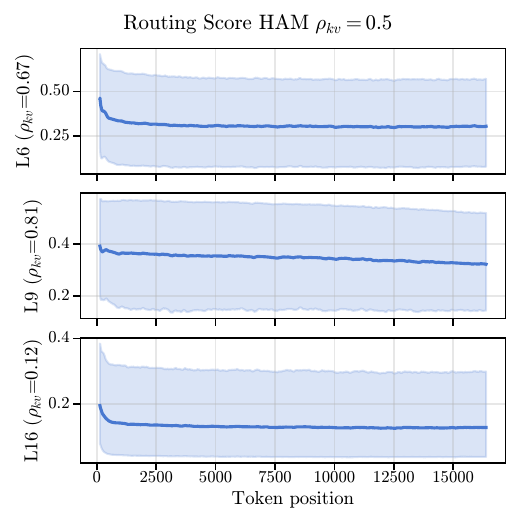}
  \caption{Routing scores $e_t$ averaged over 1,000 sequences from Institutional Books for selected layers of a HAM model with 0.5 KV usage. The average scores show a consistent downward trend along the sequence indicating that HAM is picking up long-range structure in the sequences and using its context to make better predictions towards the end of a sequence. The massive deviations around the mean trend indicate the large degree of sequence-to-sequence variability in the routing scores.
  }
  \label{fig:fit_error_trend}
\end{figure}

While the routing score is much more varied for natural language sequences, average trends over sequences are very stable.
In \cref{fig:fit_error_trend}, we analyze the average prediction error on sequences from the Institutional Books dataset \cite{cargnelutti2025institutional}, and we filter for sequences that are at least 16K tokens in length.

While there is a lot of heterogeneity from sequence to sequence, $e_t$ overall shows a downward trend indicating that the long coherent sequence get more predictable.
Different layers also have heterogeneity in their routing scores.

\begin{figure}[ht]
  \centering
  \includegraphics[width=\linewidth]{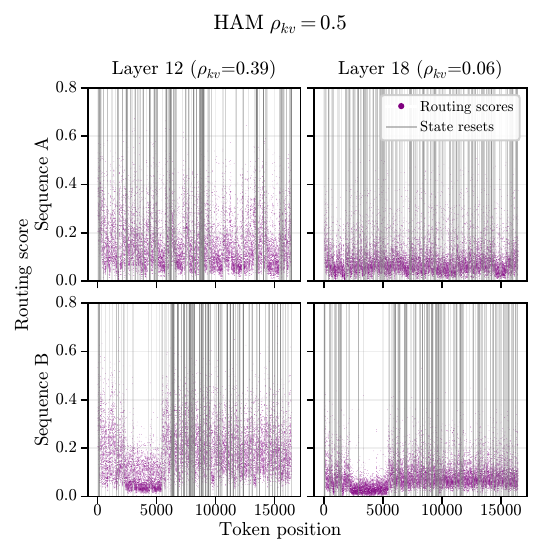}
  \caption{Routing scores and gate ($\alpha_t$) for examples sequences and layers. The vertical lines indicates time-points where $\alpha_t < 0.05$ i.e the state $S_t$ is effectively reset. In addition to the heterogeneity of the routing scores, the plot illustrates how frequently GDN resets the RNN state along the sequence.}
  \label{fig:routing_gating_examples}
\end{figure}

We want to emphasize an important point about how the routing scores interact with the gating in the gated RNNS like Gated-Deltanet.
We illustrate this with the GDN update rule:
\begin{align}
  S_t & = \alpha_t S_{t-1}\bigl(I-\beta_t k_t k_t^\top\bigr) + \beta_t v_t k_t^\top
    \label{eq:GDN_update}
\end{align}
where $\alpha_t \in [0,1]$ is an input-dependent scalar gate that controls how much $S_t$ integrates information from the past versus the current input.
In particular, when $\alpha_t$ is close to $0$, then $S_t$ is effectively reset, and the RNN starts integrating information afresh.
While adaptive integration-window is certainly powerful for performance, and from prior work on gated RNNs,  has many favourable properties \cite{yang2024gated,krishnamurthy2022theory}, it does have very important consequences for the behaviour of the routing scores: after a state reset (when $\alpha_t$ is close to $0$), the new tokens are much more likely to be surprising and thus get routed to the KV cache.
So it is important to study how $\alpha_t$ behaves along a sequence and across layers -- to our knowledge this hasn't been demonstrated in large models with real language data.
In \cref{fig:routing_gating_examples}, we show example routing scores for two sequences along with locations where the state is effectively reset ($\alpha_t < 0.05$).
As is clear the state resets are frequent and this also varies a lot with the particular sequence and across layers.
This is why the choice for the routing metric should be carefully tuned to the behaviour of the RNN on the data.
We hypothesize that this might be why the learned routers in our examples end up performing significantly better than the static prediction error routing.

\begin{figure}[ht]
  \centering
  \includegraphics[width=\linewidth]{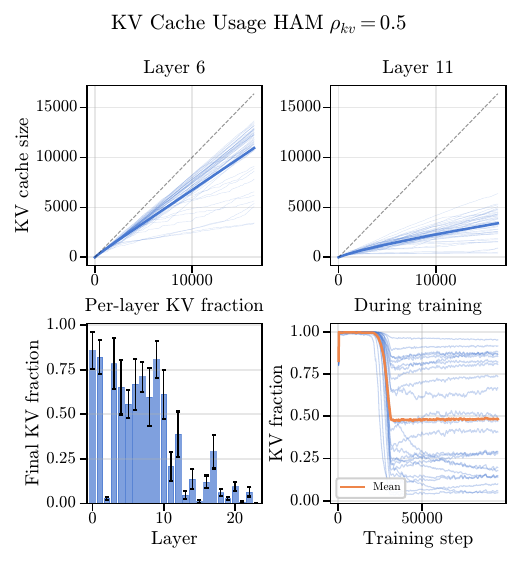}
  \caption{KV cache growth: the top panels show the mean (thick line) and some example sequence plots for the KV-cache usage for two layers on $1000$ sequences from the Institutional Books dataset. 
  The KV-cache usage for a globally set KV-cache target can be quite heterogenous across layers (bottom left), and HAM's method for tuning the KV-cache threshold quickly achieves the desired target after a period of freezing the threshold (bottom right, orange line); however, the different layers can converge to very different KV-cache usage (thin blue lines)}
  \label{fig:KV_cache_usage}
\end{figure}

\subsection{KV Cache Growth}
One of the distinctive features of HAM is the ability to control the KV cache growth in a precise manner, globally or for every layer.
This is done by the fixed or learnable threshold $\tau$ to achieve a target growth rate.
The growth rate of the KV cache determines the computational advantage of HAM over RNNs and the Transformer.
Here, we study how the KV cache grows for long coherent sequnces from the Institutional books dataset \cite{cargnelutti2025institutional}.

For most natural language data, we find that the average growth rate for the layers tends to be roughly linear, with very significant sequence-to-sequence variability.
\cref{fig:KV_cache_usage}, top panels, shows this average KV cache growth rate, averaged over $1000$ sequences from the dataset; the thin lines are per-sequence growth rates from a subset of sequences.
However, the learnable routers do show non-linear growth trends for the KV cache, which is an interesting future direction to explore (\cref{sec:KV_usage_router}).
We also see that different layers can have very different average KV cache usage when we use a global KV-cache usage target (\cref{fig:KV_cache_usage}, bottom left), suggesting that different layers might have very different requirements for the KV cache as they build richer, different representations, and HAM allows us to tune the KV usage to these requirements.
In \cref{fig:KV_cache_usage}, bottom right, we see that after a period of freezing the threshold during training, the threshold-learning algorithm (\cref{sec:arch_details}) quickly achieves the global target we set, and the individual layers still converge on very diverse KV cache usage.

A linear growth rate is advantageous for the KV cache because it prevents it from being dominated by tokens only present early in a sequence.
However, in principle, $T_\mathrm{KV}(T)$ could grow sublinearly.
Indeed, the average rate of information growth in a sequence is likely sublinear \cite{BialekNemenmanTishby2001Predictability, MerhavFeder1998UniversalPrediction}.
Insofar as surprising tokens are new information, one would hope that a properly configured complimentary KV cache would grow sublinearly as well, which would lead to subquadratic behaviour overall for the HAM model.
As discussed, in \cref{sec:ham_tradeoff}, HAM allows a very flexible, layer-wise control of the KV cache growth rate, by imposing a target schedule this growth rate can take complex forms \emph{along the sequence}; moreover, this can be done during \emph{test time}, and since the KV cache is always complemented by the RNN we do not encounter the usual problems associated with evicting tokens from the KV cache.
This gives HAM a lot of flexibility to tune the KV cache growth rate as demanded by the task at hand.

  \section{Compute and Memory Performance}


We have demonstrated that HAM can have superior computational complexity and memory usage compared to parameter-matched Transformers, depending on the size of its KV cache. 
Similarly, HAM can also have better memory usage than parameter-matched Transformers and RNNs, depending on the model size and KV cache growth rate.
\Cref{tab:model_summary_asymptotic} presents detailed expressions for how compute and memory usage scale for different model sizes, input lengths, and choices of KV cache compression rates with the HAM model.
In \cref{subsec:app-scaling}, we plot FLOPs and memory usage against different sequence lengths and model sizes, and see \cref{subsec:app-flops} for precise calculations of FLOPs and memory usage.
In \cref{tab:model_summary_asymptotic}, we summarize how parameter counts, FLOPs, and memory usage scale with hidden size $d$.



Compute and memory are not the only factors affecting HAM's performance.
A drawback of data-dependent routing to the KV cache is that it induces random access patterns from the residual stream, which can increase the chance of TLB page misses and increase latency.
This can be mitigated by modifying the routing algorithm to send contiguous blocks of tokens to the KV cache, or by increasing hidden size to exceed the page size.
Note that the access pattern for the KV cache can be determined in advance by the RNN, decoupling address generation and access, thereby greatly reducing latency, similar to DeepSeek-V3.2-Exp's Lightning Indexer \cite{deepseekai2025deepseekv32exp}.
Further, since attention computation is done in blocks, blocks can be streamed to attention and computed in parallel as soon as the first chunk of the input is processed by the RNN, minimizing latency.
For learnable routers projecting directly from the hidden state to a score, the two memories can be run fully in parallel.


  \section{Discussion}

In this work we developed a new architecture for one of the central aspects of modern LLMs---the sequence-mixing layer. 
By viewing the sequence-mixing layers as a form of associative memory we highlighted how the two most popular sequence mixing layers implement sequence-mixing in orthogonal and complementary ways in terms of computational and memory capacity on the one hand and long-context performance on the other. 
We further argued that existing hybrid models combine these two layers in a way that does not 
exploit their complementary nature. 
This led us to design a Hybrid Associative Memory layer that makes the KV cache complement the RNN memory by only storing tokens that are ``hard'' for the RNN. 
This leads to a KV cache whose growth can be controlled precisely at the level of every layer in a data-dependent way -- something that is difficult for existing architectures and post-hoc KV-cache growth control methods.
The memory/compute vs. performance in HAM behaves smoothly, thus allowing the model to select the right operating point depending on the task requirements.
The HAM model shows competitive performance with the RNN and Transfomer models. 
We used prediction (or, ``surprise'') as a criterion to route the tokens to the KV cache; however, we note that a learned criterion that learns the appropriate routing metric end-to-end works significantly better.
We emphasize that the overall framework is very flexible, and it enables users to set the KV cache usage at {\it test time} to tune compute/memory to the needs of the task.




\bibliographystyle{icml2026/icml2026}
\providecommand{\noopsort}[1]{}\providecommand{\singleletter}[1]{#1}%

  \clearpage
  \appendix
  \numberwithin{table}{section}
  \numberwithin{figure}{section}
  \onecolumn

  \section*{Appendix}

\section{Implementation and Training Details}
\label{subsec:implementation_training}

\subsection{Architectural Details}
\label{sec:arch_details}

\paragraph{Overall Architecture.}
Each HAM layer combines two complementary memory systems operating in parallel:
\begin{enumerate}
    \item \textbf{RNN Memory:} A state-based recurrent mechanism built on Gated DeltaNet (GDN), which maintains a compressed representation of all past tokens via a recurrent state matrix $\mathbf{S} \in \mathbb{R}^{d_k \times d_v}$. The state is updated with learnable decay gates and write strengths.
    \item \textbf{KV Memory:} Standard multi-head softmax attention over a \emph{dynamically selected} subset of tokens. A router determines which tokens are stored in the KV cache based on routing scores, while all tokens are processed by the RNN.
\end{enumerate}
The key innovation is \emph{complementary token selection}: at each position, a routing mechanism produces per-head scores that are compared against a learned threshold. 
Only tokens whose scores exceed the threshold are written to the KV cache, while the RNN processes every token. 
Per-head or scalar gating then combines the RNN and KV attention outputs. 
This design allows the model to allocate its finite attention budget to the tokens that are most difficult for the RNN to represent.

\paragraph{Forward Pass.}
Algorithm~\ref{alg:ham_forward} describes the forward pass of a single HAM layer during training.

\begin{algorithm}[ht]
\caption{HAM Layer Forward Pass (General Form)}\label{alg:ham_forward}
\DontPrintSemicolon
\SetAlgoLined
\SetKwComment{Comment}{$\triangleright$ }{}
\KwIn{Hidden states $\mathbf{X} \in \mathbb{R}^{T \times d}$}
\KwOut{Output hidden states $\mathbf{O} \in \mathbb{R}^{T \times d}$}
\BlankLine
\tcc{Step 1: Shared projections followed by path-specific convolutions}
$\mathbf{q},\mathbf{k},\mathbf{v} \gets \mathbf{W}_q \mathbf{X},\mathbf{W}_k \mathbf{X},\mathbf{W}_v \mathbf{X}$ \Comment*[r]{\textnormal{shared projection weights}}
$(\mathbf{q}^{\mathrm{RNN}}, \mathbf{k}^{\mathrm{RNN}}, \mathbf{v}^{\mathrm{RNN}}) \gets \mathrm{Conv}_{\mathrm{RNN}}(\mathbf{q}, \mathbf{k}, \mathbf{v})$ \Comment*[r]{\textnormal{RNN-path convolution}}
$(\mathbf{q}^{\mathrm{KV}}, \mathbf{k}^{\mathrm{KV}}, \mathbf{v}^{\mathrm{KV}}) \gets \mathrm{Conv}_{\mathrm{KV}}(\mathbf{q}, \mathbf{k}, \mathbf{v})$ \Comment*[r]{\textnormal{KV-path convolution}}
\BlankLine
\tcc{Step 2: Recurrent state update, readout, and routing}
\For{$t = 1, \dots, T$; \textnormal{each head} $h$}{
    $\mathbf{S}_t^{(h)} \gets f \left(\mathbf{S}_{t-1}^{(h)},\mathbf{k}^{\mathrm{RNN},(h)}_{t},\mathbf{v}^{\mathrm{RNN},(h)}_{t}\right)$ \Comment*[r]{\textnormal{any recurrent update}}
    $\mathbf{o}^{\mathrm{RNN},(h)}_{t} \gets F \left(\mathbf{S}_t^{(h)},\mathbf{q}^{\mathrm{RNN},(h)}_{t}\right)$ \Comment*[r]{\textnormal{associative retrieval}}
    $e_t^{(h)} \gets D \left(\mathbf{S}_{t-1}^{(h)},\mathbf{k}_t^{(h)},\mathbf{v}_t^{(h)}\right)$ \Comment*[r]{\textnormal{any routing metric}}
}
\hspace{\algorithmicindent}\textit{\small (Implemented via chunk-parallel form for training; fused recurrent form for decoding.)}\;
\BlankLine
\tcc{Step 3: Token selection via threshold $\tau$}
$m_t \gets \mathbb{I} \Big[\operatorname{Agg}_{h} \left(e_t^{(h)}\right) \geq \tau\Big]$\;
$\mathcal{C} \gets \{t : m_t = 1\}$ \Comment*[r]{\textnormal{selected cache positions}}
\BlankLine
\tcc{Step 4: Sparse attention --- all queries attend over $\mathcal{C}$}
$\mathbf{o}^{\mathrm{KV}}_{t} \gets \mathrm{Attn} \Big(\mathbf{q}^{\mathrm{KV}}_{t},\{\mathbf{k}^{\mathrm{KV}}_{i}\}_{i \in \mathcal{C}, i \leq t},\{\mathbf{v}^{\mathrm{KV}}_{i}\}_{i \in \mathcal{C}, i \leq t}\Big)$\;
\BlankLine
\tcc{Step 5: Gated combination and output}
$\mathbf{g}^{\mathrm{RNN}}_{t},\mathbf{g}^{\mathrm{KV}}_{t} \gets \mathrm{Gate}(\mathbf{X}_t)$\;
$\mathbf{O}_t \gets \mathbf{W}_o \Big(\mathbf{g}^{\mathrm{RNN}}_{t} \odot \mathrm{Norm}(\mathbf{o}^{\mathrm{RNN}}_{t}) + \mathbf{g}^{\mathrm{KV}}_{t} \odot \mathrm{Norm}(\mathbf{o}^{\mathrm{KV}}_{t})\Big)$\;
\end{algorithm}

Several design choices are worth noting:
\begin{itemize}
    \item The threshold parameter $p_\tau$ is stored in \emph{logit space} (pre-sigmoid), so the effective threshold is $\tau = \sigma(p_\tau) \in (0, 1)$.
    This ensures numerical stability and allows the AdamW optimizer to operate in an unconstrained space.
    \item The scale factor $s$ accounts for the range of the routing metric ($s = 2$ for cosine distance, $s = 1$ for learned routers).
    \item The threshold is \emph{frozen} for the first $N$ training steps (e.g., 20{,}000) to allow the model to stabilize before the routing distribution begins to adapt.
    \item The \texttt{requires\_grad} flag on $p_\tau$ is set to \texttt{False} to prevent DDP from waiting for backpropagation gradients that will never arrive; the synthetic gradient is injected manually.
\end{itemize}

In the headwise gating, two \emph{independent} sigmoid gates $\mathbf{g}_{\mathrm{KV}} = \sigma(\mathbf{W}_{\mathrm{KV}} \mathbf{x})$ and $\mathbf{g}_{\mathrm{RNN}} = \sigma(\mathbf{W}_{\mathrm{RNN}} \mathbf{x})$ multiplicatively scale the KV and RNN outputs per head, respectively.
Crucially, these are not complementary---both gates can be simultaneously high or low, allowing the model to up-weight or down-weight each memory system independently at each position.
The gated outputs are then concatenated across heads and projected to the output dimension.

\paragraph{FlexAttention and Block Masks.}
After token selection, the packed sequences have variable lengths across the batch.
We use PyTorch's \texttt{FlexAttention} with custom block masks to handle this efficiently.
The block masks enforce three constraints simultaneously:
\begin{enumerate}
    \item[(a)] \emph{Causality}---a query at original position $i$ can only attend to keys at positions $j \leq i$.
    \item[(b)] \emph{Same-document masking}---queries and keys from different documents in a packed batch cannot attend to each other.
    \item[(c)] \emph{Padding exclusion}---positions corresponding to padding tokens after selection are masked out.
\end{enumerate}
The block mask is constructed via a compiled function that maps between original token positions and packed positions.
We compile FlexAttention with \texttt{mode="max-autotune-no-cudagraphs"} for training and \texttt{mode="default"} for generation where the KV cache length changes at every step.

\paragraph{Generation.}
During autoregressive generation, the RNN state is maintained as a recurrent hidden state, while the KV cache accumulates selected tokens across decoding steps.
The block mask is cached and incrementally extended as new tokens are added.
For single-token decoding, a FlashAttention-based path is also available, which avoids FlexAttention's compilation overhead.

\paragraph{Router Implementations.}
\label{sec:router_types}
We have experimented with four router types:
\begin{enumerate}
    \item \textbf{Prediction error (\texttt{fit\_error}):} Uses the cosine distance between the GDN state prediction $\mathbf{S}_{t-1}^\top \mathbf{k}_t$ and the target value $\mathbf{v}_t$:
    \begin{equation}
        \mathrm{score}_t^{(h)} = 1 - \frac{\langle \mathbf{S}_{t-1}^{(h)\top} \mathbf{k}_t^{(h)},\mathbf{v}_t^{(h)} \rangle}{\|\mathbf{S}_{t-1}^{(h)\top} \mathbf{k}_t^{(h)}\| \cdot \|\mathbf{v}_t^{(h)}\| + \epsilon}
        \label{eq:fit_error}
    \end{equation}
    This score is computed inside the GDN Triton kernel during the recurrence loop at no additional cost.
    The score lies in $[0, 2]$, where $0$ indicates a perfect prediction and $2$ indicates anti-correlation.

    \item \textbf{Input router (\texttt{input}):} A learned function of the input hidden state $\mathbf{x}_t$.
    There are two variants of input router:
    \begin{itemize}
        \item \emph{Shallow} : A single linear projection $\mathrm{score}_t = \sigma(\mathbf{W}_r \mathbf{x}_t)$.
        \item \emph{Deep}: A 3-layer MLP with GELU activations and hidden dimension 256.
    \end{itemize}
    The input router is always evaluated in \texttt{float32} to prevent numerical collapse in \texttt{bfloat16}.


\end{enumerate}
In our experiments, we primarily use the prediction error router and the input router (both shallow and deep variants).
For the input and vector routers, routing scores are attached to the computational graph by scaling the KV values: $\mathbf{v}_{\mathrm{KV}} \gets p \cdot \mathbf{v}_{\mathrm{KV}}$, where $p = \max_h \mathrm{score}_t^{(h)}$ (or $\min_h$ depending on the aggregation mode).
This enables gradient flow through the router during backpropagation, despite the discrete token selection being non-differentiable.
The learned router uses an EDA algorithm in addtion, and this is implemented by taking the current score $e_t^{l}$ and averaging it with the previous layer's score via $e_t^{l} \gets \gamma_l e_t^{l} + (1 - \gamma_l) e_t^{l-1}$, where $\gamma_l \in (0,1)$ is initialized at 0.5 and learned for each layer.

\paragraph{Custom Triton Kernels.}
The GDN recurrence uses custom Triton kernels adapted from the Flash Linear Attention (FLA) library.
We extended the original FLA kernels to compute the fit error (Eq.~\ref{eq:fit_error}) within the recurrence loop, avoiding a separate pass over the sequence.
A chunk-based kernel with chunk size 64 is used for training (which supports the backward pass), while a fused recurrent kernel is used for short-sequence inference ($T \leq 64$).

\subsection{Training Details}
\label{sec:training_details}

\paragraph{Training Framework and Parallelism.}
We train HAM using a distributed training pipeline built on TorchTitan.
Training uses Hybrid Sharded Data Parallelism (HSDP) across 4 nodes of 8 GPUs each (32 GPUs total), \texttt{torch.compile} with dynamic shapes, and mixed-precision training in \texttt{bfloat16}.
The HSDP configuration fully shards model parameters across all data-parallel ranks, with no tensor parallelism for the 800M-scale models.

\paragraph{Optimization Hyperparameters.}
All 800M-parameter models are trained on 50 billion tokens of long-document data with a sequence length of 16{,}384 and a global batch size of $\approx$524K tokens (32 sequences per step, one per GPU).
We use AdamW with a learning rate of $10^{-4}$, $\beta_1 = 0.9$, $\beta_2 = 0.95$, $\epsilon = 10^{-15}$, and weight decay of $0.1$.
The learning rate follows a cosine decay schedule with 1{,}024 warmup steps.
Gradient norms are clipped to 1.0.
For the threshold optimizer (when learnable thresholds are enabled), we use a separate AdamW instance with learning rate $2.5 \times 10^{-4}$, $\beta_1 = 0.9$, $\beta_2 = 0.999$, and no weight decay.
Gradient norms are clipped at $1.0$. 
The vocabulary size of all models is $32{,}000$.

\paragraph{Learnable Thresholds via Synthetic Gradients.}
The routing threshold $\tau$ determines which tokens enter the KV cache.
Since token selection is a discrete, non-differentiable operation, we cannot backpropagate through it to update $\tau$.
Instead, we employ synthetic gradients that drive the actual KV cache utilization toward a user-specified target fraction $f_{\mathrm{target}}$.
Algorithm~\ref{alg:synth_grad} describes this procedure. 

\begin{algorithm}[t]
\caption{Threshold Update via Synthetic Gradients}\label{alg:synth_grad}
\DontPrintSemicolon
\SetAlgoLined
\KwIn{Threshold parameter $p_\tau \in \mathbb{R}$ (pre-sigmoid), target KV fraction $f_{\mathrm{target}}$, gain $\gamma$, clip bound $c$}

\BlankLine
\tcc{After each forward pass, on every data-parallel rank:}
\For{each sequence $i$ in the local batch}{
    $f_{\mathrm{actual}}^{(i)} \gets |\mathrm{selected\ KV\ tokens}|_i / |\mathrm{sequence\ tokens}|_i$\;
    $\mathrm{gap}^{(i)} \gets f_{\mathrm{actual}}^{(i)} - f_{\mathrm{target}}$\;
}

\BlankLine
\tcc{Aggregate across data-parallel ranks}
$\bar{g}_{\mathrm{local}} \gets \sum_i \mathrm{gap}^{(i)}$\;
$n_{\mathrm{local}} \gets |\{\mathrm{sequences\ in\ local\ batch}\}|$\;
$\bar{g} \gets \texttt{all\_reduce\_sum}(\bar{g}_{\mathrm{local}}) / \texttt{all\_reduce\_sum}(n_{\mathrm{local}})$\;

\BlankLine
\tcc{Compute synthetic gradient via sigmoid derivative (dropping $\sigma(\cdot) (1 - \sigma(\cdot))$}
$g_{\mathrm{synth}} \gets \mathrm{clamp} \left(-\gamma \cdot \bar{g},-c,c\right)$\;

\BlankLine
\tcc{Inject synthetic gradient and update with AdamW}
$p_\tau.\mathrm{grad} \mathrel{+}= g_{\mathrm{synth}}$\;
$p_\tau \gets \texttt{AdamW.step}(p_\tau)$\;
\end{algorithm}


\section{Standard LM and long-context LM tasks}
\label{sec:task-details}
We evaluate on standard zero-shot common-sense reasoning benchmarks following prior work~\citep{touvron2023llama,dao2024mamba2}. 
To evaluate models' capabilities at longer context lengths, we evaluate on RULER \cite{hsieh2024ruler}, which tests a model's ability to effectively use its entire context window, up to sequence lengths of 16K tokens. 

We report perplexity on WikiText~\citep{merity2017wikitext} and LAMBADA~\citep{paperno2016lambada}, which tests word prediction requiring broad discourse context.
For common-sense reasoning, we use HellaSwag~\citep{zellers2019hellaswag} for sentence completion about physical situations, PIQA~\citep{bisk2020piqa} for physical common-sense, ARC-Easy and ARC-Challenge~\citep{clark2018arc} for science question answering, Winogrande~\citep{sakaguchi2021winogrande} for coreference resolution requiring world knowledge, and BoolQ~\citep{clark2019boolq} for yes/no reading comprehension.

To evaluate models' capabilities at longer context lengths, we evaluate on RULER \cite{hsieh2024ruler}, which tests a model's ability to effectively use its entire context window, up to sequence lengths of 16K tokens.
RULER tests needle-in-a-haystack (NIAH) retrieval, multi-hop tracing, aggregration, and question answering over the context window.
We drop results for NIAH multikey 3 (MK3) and variable tracking (VT) because scores are uniformly low for all models at this scale; the HAM model with a learnable router is the only model to exceed single-digit accuracy, and only marginally so.

\section{Toy Example: Learnable Router}
\label{sec:app-learnable-router-toy-example}
\begin{figure}[ht]
  \centering
  \includegraphics[width=0.7\linewidth]{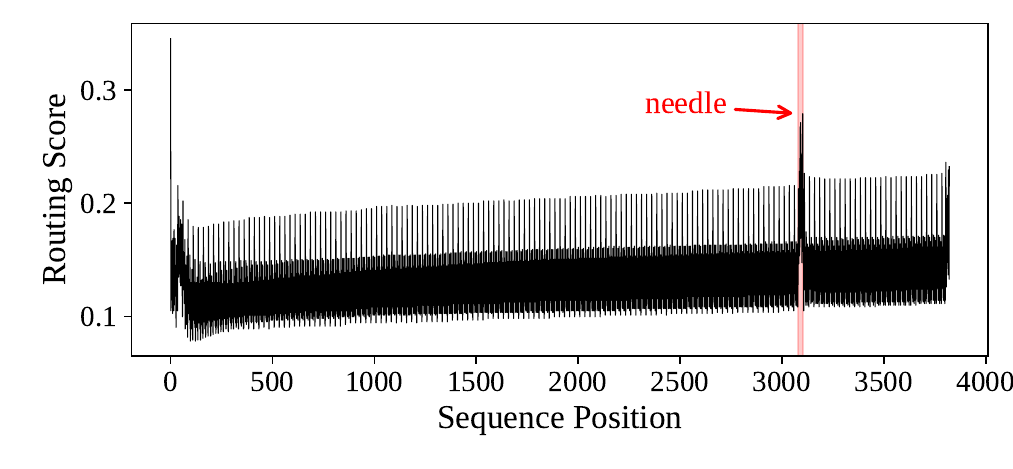}
  \caption{Trajectory of the routing for the toy NIAH single 1 sequence used in \cref{subsec:toy}.
  Again, we observe a large spike in the routing score when the needle is encountered, .
  Obtained from the 10th layer of an 800M HAM model using an end-to-end learnable router trained on 50B tokens from Long Data Collections.}
  \label{fig:toy_learnable}
\end{figure}

\section{KV cache growth rate for HAM with the learnable router}
\label{sec:KV_usage_router}
\begin{figure}[ht]
  \centering
  \includegraphics[width=0.6\linewidth]{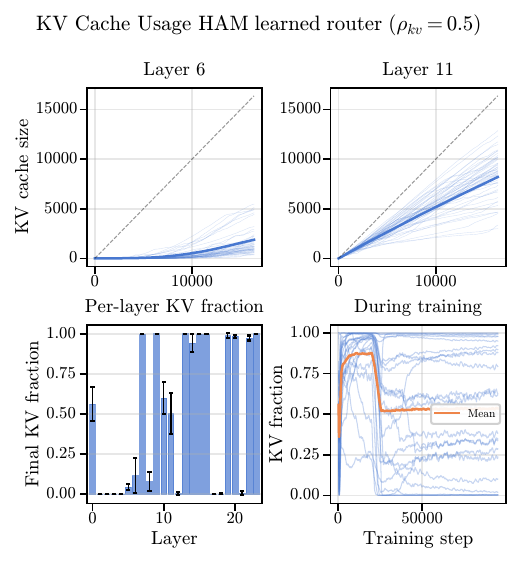}
  \caption{KV cache growth for the learnable router}
  \label{fig:KV_cache_usage-2}
\end{figure}


\section{Compute and Memory Scaling}
\label{subsec:app-scaling}

Here, we derive the compute and memory scaling for HAM, GDN, Transformer, and GDN-GSA hybrid with respect to model size, sequence length, and KV cache size.
We do this by first deriving total parameter counts and FLOPs purely as a function of the hidden size.

\subsection{Parameter Counts as a Function of Hidden Size}
In \cref{tab:params_ham,tab:params_gdn,tab:params_transformer,tab:params_ffn}, we compute the parameter counts for each HAM, GDN, Transformer, and FFN layer, respectively.
At the end of each table, we compute a ``simplified'' form that inserts relations between various configuration parameters as a function of the hidden size, $d$.
These relations are summarized below, where $d = d_{\mathrm{hidden}}$:

%
%
%

\begin{table}[ht]
\caption{Architectural relations used for simplifying parameter and FLOPs expressions.}
\label{tab:arch_relations}
\centering
\renewcommand{\arraystretch}{1.3}
\begin{footnotesize}
\begin{tabular}{l c c}
\toprule
& \textbf{HAM / GDN} & \textbf{Transformer} \\
\midrule
$d_{\mathrm{QK}}$ & $\tfrac{5}{7} d$ & $d$ \\
$d_{\mathrm{V}}$ & $\tfrac{15}{14} d$ & $d$ \\
$d_{\mathrm{RNN\text{-}QK}}^\mathrm{head}$ & $256$ & --- \\
$d_{\mathrm{RNN\text{-}V}}^\mathrm{head}$ & $384$ & --- \\
$d_{\mathrm{KV\text{-}QK}}^\mathrm{head}$ & $128$ & $128$ \\
$d_{\mathrm{KV\text{-}V}}^\mathrm{head}$ & $192$ & $128$ \\
$h_\mathrm{RNN}$ & $d_\mathrm{QK} / d_\mathrm{RNN\text{-}QK}^\mathrm{head}$ & --- \\
$h_\mathrm{KV}$ & $d_\mathrm{QK} / d_\mathrm{KV\text{-}QK}^\mathrm{head}$ & $d / d_\mathrm{KV\text{-}QK}^\mathrm{head}$ \\
FFN hidden ratio $r$ & $\tfrac{15}{7}$ & $2$ \\
\bottomrule
\end{tabular}
\end{footnotesize}
\end{table}

\noindent For HAM, we further assume no learnable thresholds and no learnable router (i.e., routing via the cosine prediction error metric).

First, we derive the total parameter count for each model as a function of $d$ and the number of layers.
Let $L_\mathrm{HAM}$, $L_\mathrm{GDN}$, and $L_\mathrm{TF}$ denote the number of HAM, GDN, and Transformer layers, respectively, and $V = 32,000$ be the vocabulary size.
Then, from \cref{tab:params_ham,tab:params_gdn,tab:params_transformer,tab:params_ffn}, we have
\begin{align}
	\mathcal{P}_\mathrm{HAM}
	&= L_\mathrm{HAM} \left[ \left( \frac{8345}{1792} d^2+\frac{23301}{896} d+576 \right) + \left( \frac{30}{7} d^2 + d \right) \right] + (2 V + 1) d \\
	&= L_\mathrm{HAM} \left[ \frac{16025}{1792} d^2 + \frac{24197}{896} d + 576 \right] + 64001 d \\
	\mathcal{P}_\mathrm{GDN}
	&= L_\mathrm{GDN} \left[ \left( \frac{65}{14}d^{2}+21d+394 \right) + \left( \frac{30}{7} d^2 + d \right) \right] + (2 V + 1) d \\
	&= L_\mathrm{GDN} \left[ \frac{125}{14} d^2 + 22d + 394 \right] + 64001 d \\
	\mathcal{P}_\mathrm{TF}
	&= L_\mathrm{TF} \left[ d(4d+1) + d(4d+1) \right] + (2 V + 1) d \\
	&= L_\mathrm{TF} \left[ 2d ( 4d + 1) \right] + 64001 d
\end{align}

We consider a scaling regime that fixes the aspect ratio (ratio of hidden size to the number of layers).
We define aspect ratios $rho$ for each model using the 800M configs used for all experimental results in this work: $\rho_\mathrm{HAM} = \rho_\mathrm{GDN} = \rho_\mathrm{GDN\text{-}GSA} = \frac{1792}{24} \approx 74.67$ and $\rho_\mathrm{TF} = \frac{1920}{23} \approx 83.48$.
We use this to obtain parameter counts strictly as a function of $d$:
\begin{align}
	\mathcal{P}_\mathrm{HAM}
	&= \frac{d}{\rho_\mathrm{HAM}} \left[ \frac{16025}{1792} d^2 + \frac{24197}{896} d + 576 \right] + 64001 d \\
	&= \frac{48075}{401408}d^3 + \frac{72591}{200704} d^2 + \frac{448061}{7} d \\
	\mathcal{P}_\mathrm{GDN}
	&= \frac{d}{\rho_\mathrm{GDN}} \left[ \frac{125}{14} d^2 + 22d + 394 \right] + 64001 d \\
	&= \frac{375}{3136} d^3 + \frac{33}{112} d^2 + \frac{7168703}{112} d \\
	\mathcal{P}_\mathrm{TF}
	&= \frac{d}{\rho_\mathrm{TF}} \left[ 2d ( 4d + 1) \right] + 64001 d \\
	&= \frac{23}{240} d^3 + \frac{23}{960} d^2 + 64001 d.
\end{align}
We can similarly derive the parameter counts for the GDN-GSA hybrid, which alternates GDN and Transformer layers.
We consider the more general case where every $k$-th layer is GSA (assuming that $k$ divides the number of layers:
\begin{align}
	\mathcal{P}_\mathrm{GDN\text{-}GSA}^{k}
	&= \underbrace{ \left( 1 - \frac{1}{k} \right) L_\mathrm{GDN\text{-}GSA} \left[ \frac{125}{14} d^2 + 22d + 394 \right] }_\text{GDN layers}
	+  \underbrace{ \frac{L_\mathrm{GDN\text{-}GSA}}{k} \left[ d(4d+1) + \left( \frac{30}{7} d^2 + d \right) \right] }_\text{GSA layers} + (2V+1) d \\
	&= \left(\frac{375k-27}{3136k}\right)d^3 + \left(\frac{33k-30}{112k}\right)d^2 + \left(\frac{7168703k-591}{112k}\right)d.
\end{align}

\subsection{Forward FLOPs as a Function of Hidden Size}
In \cref{tab:flops_ham,tab:flops_gdn,tab:flops_transformer,tab:flops_ffn}, we compute the forward FLOPs for these layers, reusing the relations from above for parameter counts to derive simplifications.
We derive the total forward FLOPs for each model in terms of $d$ using the aspect ratios from above:
\begin{align}
	\mathcal{F}_\mathrm{HAM}
	&= T L_\mathrm{HAM} \left[ 
	\left( \frac{65}{7}d^2 + \frac{33059}{14}d + 13669 + T_{\mathrm{KV}}\left(\frac{25}{14}d + 20\right) \right) + \frac{20}{7} d (3d + 2) 
	 \right] + 4 T V d + 4 T d \\
	 &= T \left[ \frac{375}{1568}d^{3}+\frac{99417}{3136}d^{2}+\frac{28713903}{224}d \right]
	 + T T_{KV} \left[ \frac{75}{3136}d^{2}+\frac{15}{56}d \right] \\
	 \mathcal{F}_\mathrm{GDN}
	 &= T L_\mathrm{GDN} \left[ 
	 d \left( \frac{2085}{224}d+\frac{1560037}{1792}\right) + \frac{20}{7} d (3d + 2) 
	 \right] + 4 T V d + 4 T d \\
	 &= T \left[ \frac{12015}{50176} d^3 + \frac{4689327}{401408}d^2 + 128004d \right] \\
	 \mathcal{F}_\mathrm{TF}
	 &= T L_\mathrm{TF} \left[ 
	 8d (d+2) + \frac{129}{64} T d + \frac{8}{3} d (3d + 2)
	 \right] + 4 T V d + 4 T d \\
	 &= T\left[ \frac{23}{120}d^3 + \left( \frac{23}{90} + \frac{989}{40960}T \right) d^2 + 128004d \right] \\
	 \mathcal{F}_\mathrm{GDN\text{-}GSA}^k
	 &= T \left( 1 - \frac{1}{k} \right) L_\mathrm{GDN\text{-}GSA} \left[ d \left( \frac{2085}{224}d+\frac{1560037}{1792}\right) + \frac{20}{7} d (3d + 2) \right] \\
	 &+ T \frac{L_\mathrm{GDN\text{-}GSA}}{k} \left[ 8d (d+2) + \frac{129}{64} T d + \frac{20}{7} d (3d + 2) \right]
	 + 4VT d + 4Td \\
	 &= T \left[ \left(\frac{12015k-879}{50176k}\right)d^{3}+\left(\frac{10836T+4689327k-4572591}{401408k}\right)d^{2}+128004d \right].
\end{align}

\subsection{Training FLOPs}
\label{subsubsec:training_flops}
We can use the results of the previous section to compute the training FLOPs for the models presented in the main text.
We use the simple heuristic that training FLOPs $\approx 3$ $\times$ forward FLOPs.
The total forward FLOPs during training are:
\begin{align}
	\mathcal{F}(T=16384)/\text{rank}/\text{step} \times \text{32 ranks } \times (S = \text{95367 steps}).
\end{align}
\Cref{tab:training_flops} shows the training FLOPs for each model we run experiments for.

\begin{table}[ht]
\caption{Empirical training FLOPs for each model.}
\label{tab:training_flops}
\centering
\renewcommand{\arraystretch}{1.4}
\begin{footnotesize}
\begin{tabular*}{0.85\textwidth}{l @{\extracolsep{\fill}} r @{\hspace{6pt}} r}
\toprule
\textbf{Model} & \textbf{Training FLOPs (zFLOPs)} & \textbf{Relative} \\
\midrule
HAM ($T_\mathrm{KV}^*/T = 0.5$) & 0.3511 & \emph{ref.} \\
GDN & 0.2467 & $-$29.7\% \\
Transformer & 0.4592 & +30.8\% \\
GDN-GSA ($k=2$) & 0.3429 & $-$2.3\% \\
\bottomrule
\end{tabular*}
\end{footnotesize}
\end{table}

\subsection{Forward Memory}
Here, we calculate the memory required for forward passes through each model.
We consider the memory needed to store weights and to cache associative memory states (the GDN state and KV cache), but ignore lower-order terms like 1D-convolution states.
In BF16, each model requires $2 \mathcal{P}$ bytes for the weights alone.
The GDN state, which is also stored in BF16, has size $2 L_\mathrm{RNN} d_\mathrm{QK} d_\mathrm{RNN\text{-}V}$, where $L$ is the number of RNN layers.
The KV cache, also in BF16, has size $2 L_\mathrm{KV} T_\mathrm{KV} (d_\mathrm{QK} + d_\mathrm{V})$.
From this, we can derive the forward memory for each model.
\begin{align}
	\mathcal{M}_\mathrm{HAM}
	&= 2\mathcal{P}_\mathrm{HAM} + 2L_\mathrm{HAM} \left( d_\mathrm{QK} d_\mathrm{RNN\text{-}V}^\mathrm{head} + T_\mathrm{KV} (d_\mathrm{QK} + d_\mathrm{V}) \right) \\
	&= \frac{48075}{200704} d^3 + \frac{809871}{100352} d^2 + \frac{75}{1568} T_\mathrm{KV} d^2 + \frac{896122}{7} d \\
	\mathcal{M}_\mathrm{GDN}
	&= 2\mathcal{P}_\mathrm{GDN} + 2L_\mathrm{GDN} d_\mathrm{QK} d_\mathrm{RNN\text{-}V}^\mathrm{head} \\
	&= \frac{375}{1568} d^3 + \frac{3111}{392} d^2 + \frac{7168703}{56} d \\
	\mathcal{M}_\mathrm{TF}
	&= 2\mathcal{P}_\mathrm{TF} + 2L_\mathrm{TF} T (d_\mathrm{QK} + d_\mathrm{V}) \\
	&= \frac{23}{120} d^3 + \frac{23}{480} (1+T) d^2 + 128002 d \\
	\mathcal{M}_\mathrm{GDN\text{-}GSA}^k
	&= 2\mathcal{P}_\mathrm{GDN\text{-}GSA}^k + 2L_\mathrm{GDN\text{-}GSA} \left[ \tfrac{k-1}{k} d_\mathrm{QK} d_\mathrm{RNN\text{-}V}^\mathrm{head} + \tfrac{T}{k} (d_\mathrm{QK} + d_\mathrm{V}) \right] \\
	&= \left(\frac{375k-27}{1568k}\right) d^3 + \left(\frac{12444k-12360+75T}{1568k}\right) d^2 + \left(\frac{7168703k-591}{56k}\right) d
\end{align}

\subsection{Parameters vs. Forward FLOPs and Memory}
\begin{figure}[ht]
  \centering
  \includegraphics[width=0.3\textwidth]{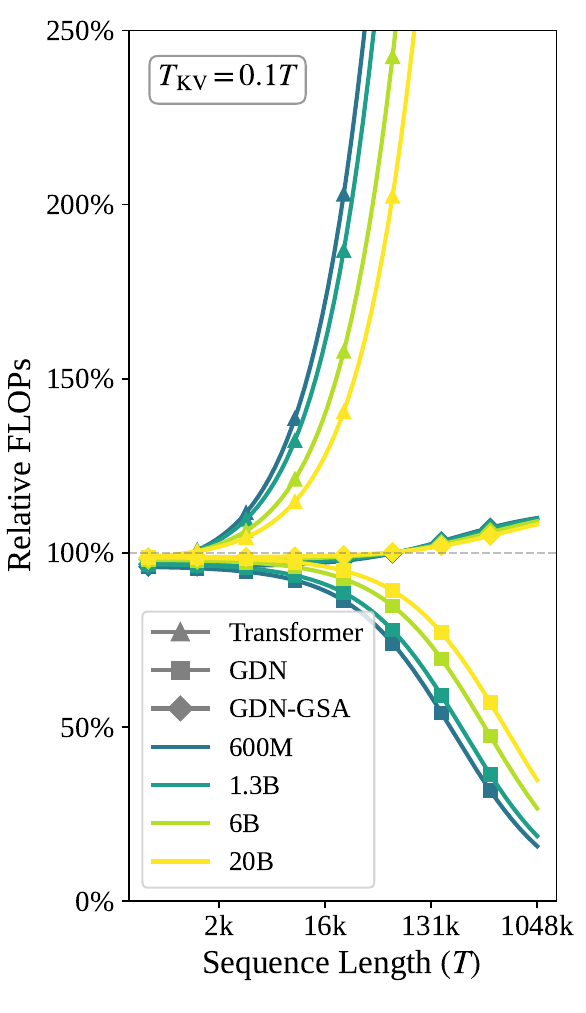}
  \includegraphics[width=0.3\textwidth]{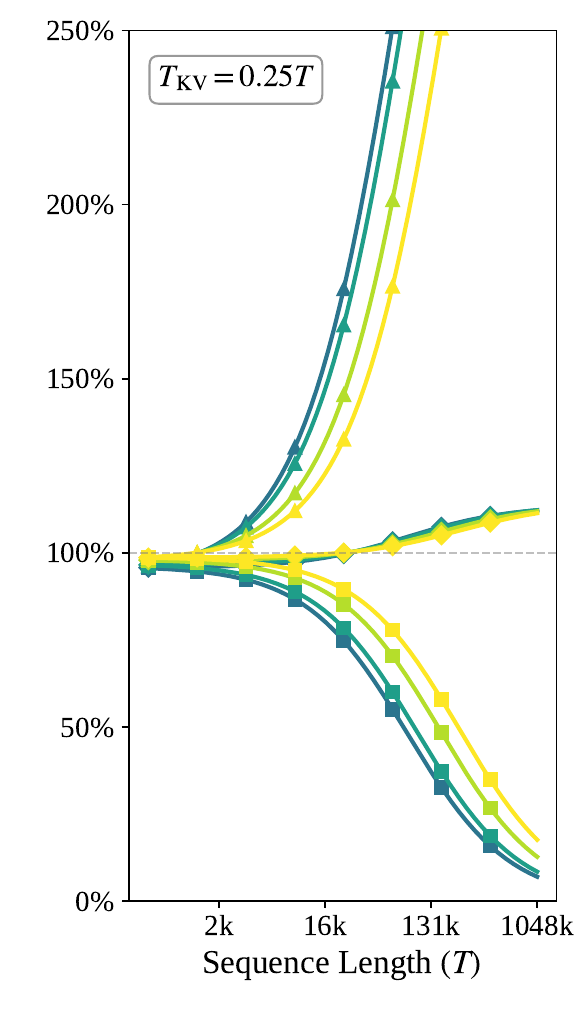}
  \includegraphics[width=0.3\textwidth]{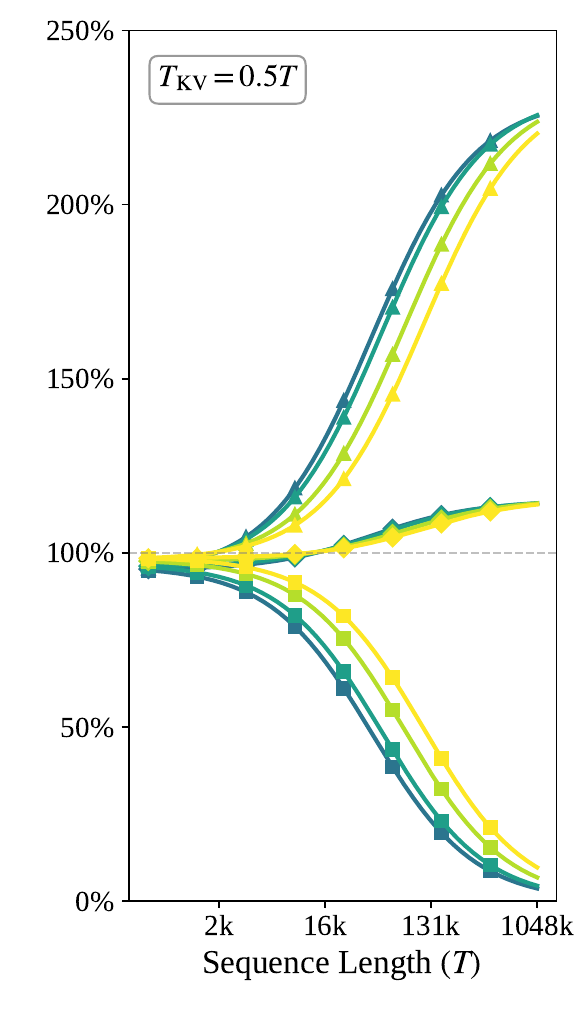}
  \caption{Forward FLOPs for GDN, Transformer, and GDN-GSA relative to HAM, for $T_\mathrm{KV}/T = 0.10$ (left), $T_\mathrm{KV}/T = 0.25$ (center), and $T_\mathrm{KV}/T = 0.5$ (right).}
  \label{fig:flops-ratio}
\end{figure}

\begin{figure}[ht]
  \centering
  \includegraphics[width=0.3\textwidth]{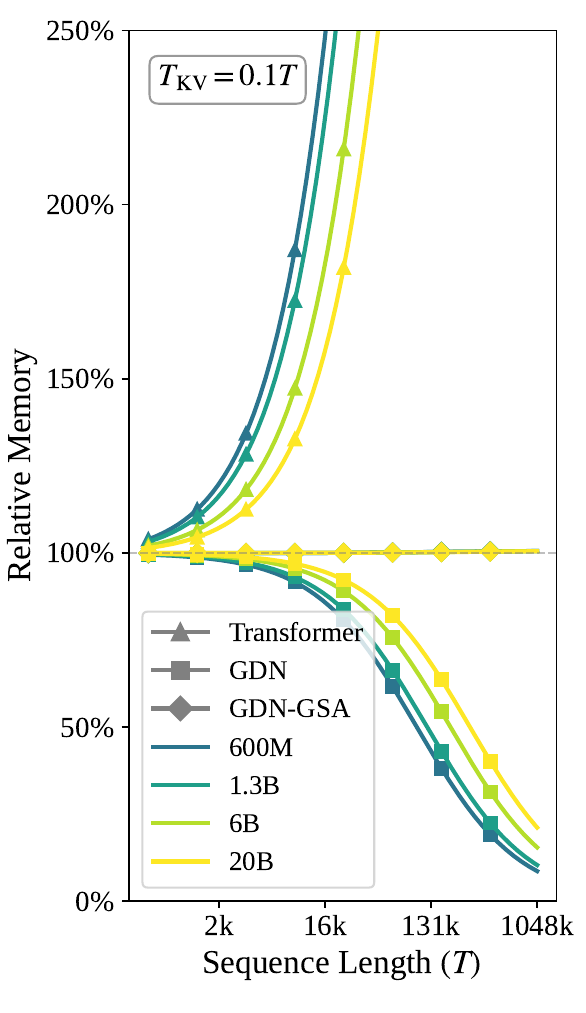}
  \includegraphics[width=0.3\textwidth]{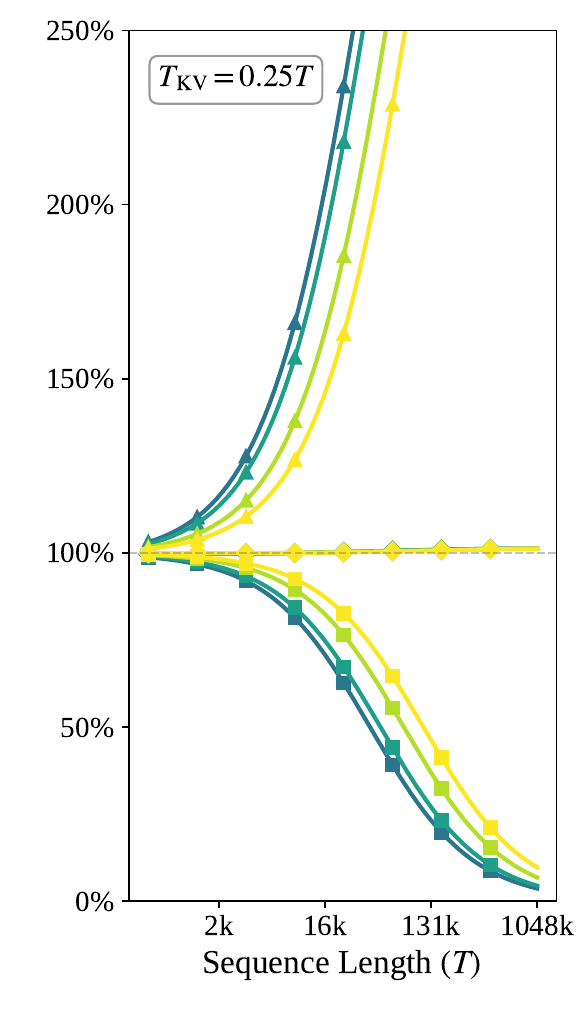}
  \includegraphics[width=0.3\textwidth]{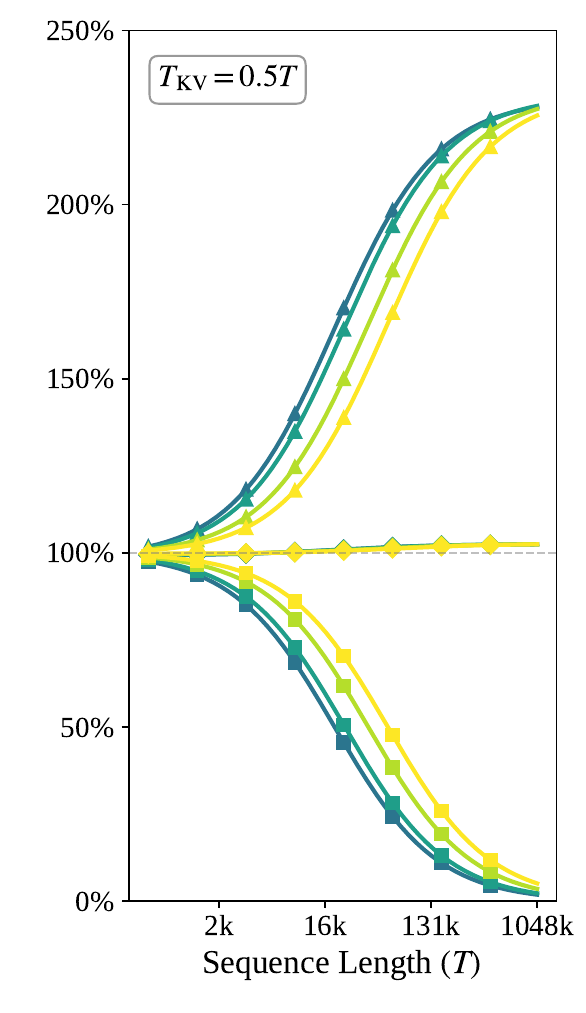}
  \caption{Forward memory for GDN, Transformer, and GDN-GSA relative to HAM, for $T_\mathrm{KV}/T = 0.10$ (left), $T_\mathrm{KV}/T = 0.25$ (center), and $T_\mathrm{KV}/T = 0.5$ (right).}
  \label{fig:memory-ratio}
\end{figure}

In \cref{tab:params_scaling,tab:flops_scaling} we summarize our calculations above.
From \cref{tab:params_scaling}, we find the hidden size $d$ yielding a desired model size, $\mathcal{P}$, and compute its FLOPs function, $\mathcal{F}(T)$.
We use this to generate a suite of relative FLOPs and memory curves for the models in \cref{fig:flops-ratio,fig:memory-ratio}.

We can also derive simplified expressions for $\mathcal{F}$ and $\mathcal{M}$ for each model.
We do not show all the steps here, but we use Cardano's formula to obtain $\mathcal{P}(d)$, plug into the FLOPs/memory expression, and drop lower-order terms in $P$ and $T$ until we obtain a simple yet faithful approximation.
We present these approximations in \cref{tab:model_summary_asymptotic}.

\begin{table}[ht]
\caption{Total model parameters as a function of $d$.}
\label{tab:params_scaling}
\centering
\renewcommand{\arraystretch}{1.4}
\begin{footnotesize}
\begin{tabular*}{0.85\textwidth}{l @{\extracolsep{\fill}} c}
\toprule
\textbf{Model} & \textbf{Parameters} \\
\midrule
HAM & $\frac{48075}{401408}d^3 + \frac{72591}{200704} d^2 + \frac{448061}{7} d$ \\
GDN & $\frac{375}{3136} d^3 + \frac{33}{112} d^2 + \frac{7168703}{112} d$ \\
Transformer & $\frac{23}{240} d^3 + \frac{23}{960} d^2 + 64001 d$ \\
GDN-GSA & $\left(\frac{375k-27}{3136k}\right)d^3 + \left(\frac{33k-30}{112k}\right)d^2 + \left(\frac{7168703k-591}{112k}\right)d$ \\
\bottomrule
\end{tabular*}
\end{footnotesize}
\end{table}

\begin{table}[ht]
\caption{Total forward FLOPs as a function of $d$.}
\label{tab:flops_scaling}
\centering
\renewcommand{\arraystretch}{1.4}
\begin{footnotesize}
\begin{tabular*}{0.85\textwidth}{l @{\extracolsep{\fill}} c}
\toprule
\textbf{Model} & \textbf{Forward FLOPs} \\
\midrule
HAM & $T \left[ \frac{375}{1568}d^{3}+\frac{99417}{3136}d^{2}+\frac{28713903}{224}d \right] + T T_{\mathrm{KV}} \left[ \frac{75}{3136}d^{2}+\frac{15}{56}d \right]$ \\
GDN & $T \left[ \frac{12015}{50176} d^3 + \frac{4689327}{401408}d^2 + 128004d \right]$ \\
Transformer & $T\left[ \frac{23}{120}d^3 + \left( \frac{23}{90} + \frac{989}{40960}T \right) d^2 + 128004d \right]$ \\
GDN-GSA & $T \left[ \left(\frac{12015k-879}{50176k}\right)d^{3}+\left(\frac{10836T+4689327k-4572591}{401408k}\right)d^{2}+128004d \right]$ \\
\bottomrule
\end{tabular*}
\end{footnotesize}
\end{table}

\begin{table}[ht]
\caption{Total forward memory as a function of $d$ (bytes, BF16).}
\label{tab:memory_scaling}
\centering
\renewcommand{\arraystretch}{1.3}
\begin{footnotesize}
\begin{tabular*}{0.85\textwidth}{l @{\extracolsep{\fill}} c}
\toprule
\textbf{Model} & \textbf{Forward Memory} (bytes) \\
\midrule
HAM & $\frac{48075}{200704} d^3 + \frac{809871}{100352} d^2 + \frac{75}{1568} T_\mathrm{KV} d^2 + \frac{896122}{7} d$ \\
GDN & $\frac{375}{1568}d^3+\frac{3111}{392}d^2+\frac{7168703}{56}d$ \\
Transformer & $\frac{23}{120} d^3 + \frac{23}{480} (1+T) d^2 + 128002 d$ \\
GDN-GSA & $\left(\frac{375k-27}{1568k}\right) d^3 + \left(\frac{12444k-12360+75T}{1568k}\right) d^2 + \left(\frac{7168703k-591}{56k}\right) d$ \\
\bottomrule
\end{tabular*}
\end{footnotesize}
\end{table}

\begin{table*}[ht]
\caption{Approximate scaling for longer contexts.
All expressions are accurate to within ${\sim}2\%$ of the full expressions in \cref{tab:model_summary_asymptotic}.}
\label{tab:model_summary_asymptotic}
\vskip 0.15in
\begin{center}
\begin{footnotesize}
\begin{tabular*}{0.85\textwidth}{l @{\extracolsep{\fill}} l l}
\toprule
\textbf{Model} & \textbf{Forward FLOPs} (per token) & \textbf{Forward Memory} (bytes) \\
\midrule
HAM & $2P+(130+0.1 T_{KV})P^{\frac{2}{3}}-8.5\text{K} T_{KV}$ & $2P+T_{KV}(0.208 P^{\frac{2}{3}}-13 P^{\frac{1}{3}}-11.4\text{K})$ \\
GDN & $2P+46 P^{\frac{2}{3}}$ & $2P+30 P^{\frac{2}{3}}$ \\
Transformer & $2P+T(0.115 P^{\frac{2}{3}}-11\text{K})$ & $2P+T(0.23 P^{\frac{2}{3}}-21\text{K})$ \\
GDN-GSA ($k=2$) & $2P+T(0.06 P^{\frac{2}{3}}-4 P^{\frac{1}{3}}-3.3\text{K})$ & $2P+T(0.107 P^{\frac{2}{3}}-7 P^{\frac{1}{3}}-5.9\text{K})$ \\
\bottomrule
\end{tabular*}
\end{footnotesize}
\end{center}
\vskip -0.1in
\end{table*}

\label{subsec:app-params}

%

\begin{table*}[ht]
\caption{Per-layer parameter count for HAM.}
\label{tab:params_ham}
\centering
\renewcommand{\arraystretch}{1.2}

\newcommand{\colgap}{\hspace{64pt}}

\begin{small}
\begin{tabular*}{\textwidth}{l @{\extracolsep{\fill}} r}
\hline
\textbf{Source of weights} & \textbf{\# parameters} \\
\hline

\multicolumn{2}{l}{\textbf{Pre-norm}} \\
Pre-norm
& $d_\mathrm{hidden}$ \\

\emph{Subtotal}
& $d_\mathrm{hidden}$ \\
\hline

\multicolumn{2}{l}{\textbf{Q, K, V projections} (shared)} \\

$W_{\mathrm{Q}}, W_{\mathrm{K}}$ 
& $2 \cdot d_\mathrm{hidden} \cdot d_{\mathrm{QK}}$ \\

$W_{\mathrm{V}}$ 
& $d_\mathrm{hidden} \cdot d_{\mathrm{V}}$ \\

%

\emph{Subtotal}
& $d_\mathrm{hidden} (2 d_{\mathrm{QK}} + d_{\mathrm{V}} )$ \\
\hline

\multicolumn{2}{l}{\textbf{RMS norms for Q, K, V} (separate)} \\

RNN RMS norms
& $2 \cdot d_{\mathrm{QK}} + d_{\mathrm{V}}$ \\

KV RMS norms
& $2 \cdot d_{\mathrm{QK}} + d_{\mathrm{V}}$ \\

\emph{Subtotal}
& $2 (2 d_{\mathrm{QK}} + d_{\mathrm{V}})$ \\
\hline

\multicolumn{2}{l}{\textbf{RNN scalars (from GDN)}} \\

$a, b$ projections
& $2 \cdot d_\mathrm{hidden} \cdot h_{\mathrm{RNN}}$ \\

$A_{\log}$
& $h_{\mathrm{RNN}}$ \\

$dt$ bias
& $h_{\mathrm{RNN}}$ \\

\emph{Subtotal}
& $2 h_{\mathrm{RNN}} (d_\mathrm{hidden} + 1)$ \\
\hline

\multicolumn{2}{l}{\textbf{1D convolutions} (separate)} \\

$W_{\mathrm{RNN\text{-}Q}}^{\mathrm{conv}}, W_{\mathrm{RNN\text{-}K}}^{\mathrm{conv}}$ 
& $2 \cdot d_{\mathrm{QK}} \cdot d_{\mathrm{conv}}$ \\

$W_{\mathrm{RNN\text{-}V}}^{\mathrm{conv}}$ 
& $d_{\mathrm{V}} \cdot d_{\mathrm{conv}}$ \\

$W_{\mathrm{KV\text{-}Q}}^{\mathrm{conv}}, W_{\mathrm{K}}^{\mathrm{conv}}$ 
& $2 \cdot d_{\mathrm{QK}} \cdot d_{\mathrm{conv}}$ \\

$W_{\mathrm{V}}^{\mathrm{conv}}$ 
& $d_{\mathrm{V}} \cdot d_{\mathrm{conv}}$ \\

\emph{Subtotal}
& $2 d_{\mathrm{conv}} (2 d_{\mathrm{QK}} + d_{\mathrm{V}})$ \\
\hline

\multicolumn{2}{l}{\textbf{KV token selection}} \\

$\tau_\ell$  (if using learnable thresholds)
& $1$ \\

$W_\mathrm{router}$ (if using a learnable router)
& $d_\mathrm{hidden}$ \\

\emph{Subtotal}
& $\mathbbm{1}(\text{learnable $\tau_\ell$}) + d_\mathrm{hidden} \cdot \mathbbm{1}(\text{learnable router})$ \\
\hline

\multicolumn{2}{l}{\textbf{RNN \& KV headwise output normalization} (separate)} \\

RNN norm
& $d_{\mathrm{RNN\text{-}V}}^{\mathrm{head}}$ \\

RNN norm gate
& $d_{\mathrm{hidden}} \cdot d_{\mathrm{V}}$ \\

KV norm
& $d_{\mathrm{KV\text{-}V}}^{\mathrm{head}}$ \\

\emph{Subtotal}
& $d_{\mathrm{RNN\text{-}V}}^{\mathrm{head}} + d_{\mathrm{KV\text{-}V}}^{\mathrm{head}} + d_{\mathrm{hidden}} \cdot d_{\mathrm{V}}$ \\
\hline

\multicolumn{2}{l}{\textbf{RNN \& KV gating}} \\


Per-head RNN gate
& $d_\mathrm{hidden} \cdot h_{\mathrm{RNN}}$ \\

Per-head KV gate
& $d_\mathrm{hidden} \cdot h_{\mathrm{KV}}$ \\


\emph{Subtotal}
& $d_\mathrm{hidden} (h_{\mathrm{RNN}} + h_{\mathrm{KV}})$ \\
\hline

%
%
%
%

\multicolumn{2}{l}{\textbf{Output projection}} \\

Output projection
& $d_\mathrm{V} \cdot d_\mathrm{hidden}$ \\


\emph{Subtotal}
& $d_\mathrm{V} d_\mathrm{hidden}$ \\

\noalign{\hrule height 1pt}

\textbf{Layer Total}

& $(d_\mathrm{hidden}+2d_\mathrm{conv}+2)(2d_\mathrm{QK}+d_\mathrm{V})
+ d_\mathrm{hidden}(3h_\mathrm{RNN}+h_\mathrm{KV}+2d_\mathrm{V}+1)
+ 2h_\mathrm{RNN}$ \\
& $+\, d_{\mathrm{RNN\text{-}V}}^{\mathrm{head}}
+ d_{\mathrm{KV\text{-}V}}^{\mathrm{head}}
+ \mathbbm{1}(\text{learnable }\tau_\ell)
+ d_\mathrm{hidden}\cdot\mathbbm{1}(\text{learnable router})$ \\


\emph{Simplified}

& $\frac{8345}{1792} d^2+\frac{23301}{896} d+576$ \\
\hline

\end{tabular*}
\end{small}

\end{table*}

%

\begin{table*}[ht]
\caption{Per-layer parameter count for Transformer.}
\label{tab:params_transformer}
\centering
\renewcommand{\arraystretch}{1.2}

\newcommand{\colgap}{\hspace{64pt}}

\begin{small}
\begin{tabular*}{\textwidth}{l @{\extracolsep{\fill}} r}
\hline
\textbf{Source of weights} & \textbf{\# parameters} \\
\hline

\multicolumn{2}{l}{\textbf{Pre-norm}} \\
Pre-norm
& $d_\text{hidden}$ \\

\emph{Subtotal}
& $d_\text{hidden}$ \\
\hline

\multicolumn{2}{l}{\textbf{Q, K, V projections}} \\

$W_{\text{Q}}$
& $d_{\text{hidden}} \cdot d_{\text{hidden}}$ \\

$W_{\text{K}}$
& $d_{\text{hidden}} \cdot d_{\mathrm{KV}}$ \\

$W_{\mathrm{V}}$
& $d_{\text{hidden}} \cdot d_{\mathrm{KV}}$ \\

\emph{Subtotal}
& $d_{\text{hidden}} (d_{\text{hidden}} + 2 d_{\mathrm{KV}})$ \\
\hline

%
%
%

\multicolumn{2}{l}{\textbf{Output projection}} \\

$W_{\text{O}}$
& $d_{\text{hidden}} \cdot d_{\text{hidden}}$ \\

\emph{Subtotal}
& $d_{\text{hidden}}^2$ \\

\noalign{\hrule height 1pt}
\textbf{Layer Total}
& $d_{\text{hidden}} (2 d_{\text{hidden}} + 2 d_{\mathrm{KV}} + 1)$ \\

\emph{Simplified}
& $d(4d+1)$ \\
\hline

\end{tabular*}
\end{small}

\end{table*}

%

\begin{table*}[ht]
\caption{Per-layer parameter count for Gated DeltaNet.}
\label{tab:params_gdn}
\centering
\renewcommand{\arraystretch}{1.2}

\newcommand{\colgapgdn}{\hspace{64pt}}

\begin{small}
\begin{tabular*}{\textwidth}{l @{\extracolsep{\fill}} r}
\hline
\textbf{Source of weights} & \textbf{\# parameters} \\
\hline

\multicolumn{2}{l}{\textbf{Pre-norm}} \\
Pre-norm
& $d_\text{hidden}$ \\

\emph{Subtotal}
& $d_\text{hidden}$ \\
\hline

\multicolumn{2}{l}{\textbf{Q, K, V projections}} \\

$W_{\text{Q}}, W_{\text{K}}$
& $2 \cdot d_{\text{hidden}} \cdot d_{\mathrm{QK}}$ \\

$W_{\mathrm{V}}$
& $d_{\text{hidden}} \cdot d_{\mathrm{V}}$ \\

\emph{Subtotal}
& $d_{\text{hidden}} (2 d_{\mathrm{QK}} + d_{\mathrm{V}})$ \\
\hline

\multicolumn{2}{l}{\textbf{RNN scalars}} \\

$a, b$ projections
& $2 \cdot d_{\text{hidden}} \cdot h$ \\

$A_{\log}$
& $h$ \\

$dt$ bias
& $h$ \\

\emph{Subtotal}
& $2 h (d_{\text{hidden}} + 1)$ \\
\hline

\multicolumn{2}{l}{\textbf{1D convolutions}} \\

$W_{\text{Q}}^{\text{conv}}, W_{\text{K}}^{\text{conv}}$
& $2 \cdot d_{\mathrm{QK}} \cdot d_{\text{conv}}$ \\

$W_{\mathrm{V}}^{\text{conv}}$
& $d_{\mathrm{V}} \cdot d_{\text{conv}}$ \\

\emph{Subtotal}
& $d_{\text{conv}} (2 d_{\mathrm{QK}} + d_{\mathrm{V}})$ \\
\hline

\multicolumn{2}{l}{\textbf{Output gating}} \\

Gate projection 
& $d_{\text{hidden}} \cdot d_{\mathrm{V}}$ \\

Output norm 
& $d_{\mathrm{V}}^{\text{head}}$ \\

\emph{Subtotal}
& $d_{\text{hidden}} d_{\mathrm{V}} + d_{\mathrm{V}}^{\text{head}}$ \\
\hline

\multicolumn{2}{l}{\textbf{Output projection}} \\

$W_{\text{O}}$
& $d_{\mathrm{V}} \cdot d_{\text{hidden}}$ \\

\emph{Subtotal}
& $d_{\mathrm{V}} d_{\text{hidden}}$ \\

\noalign{\hrule height 1pt}
\textbf{Layer Total}
& $d_{\text{hidden}} (2 d_{\mathrm{QK}} + 3 d_{\mathrm{V}} + 2h + 1) + d_{\text{conv}} (2 d_{\mathrm{QK}} + d_{\mathrm{V}}) + 2h +  d_{\mathrm{V}}^{\text{head}}$ \\

\emph{Simplified}
& $\frac{65}{14}d^{2}+21d+394$ \\


\hline

\end{tabular*}
\end{small}

\end{table*}

%

\begin{table}[ht]
\caption{Per-layer parameter count for FFN (SwiGLU).}
\label{tab:params_ffn}
\centering
\renewcommand{\arraystretch}{1.2}

\newcommand{\colgap}{\hspace{64pt}}

\begin{small}
\begin{tabular*}{\textwidth}{l @{\extracolsep{\fill}} r}
\hline
\textbf{Source of weights} & \textbf{\# parameters} \\
\hline

\multicolumn{2}{l}{\textbf{FFN projections}} \\

Pre-norm
& $d_\text{hidden}$ \\

Gate projection ($W_{\text{gate}}$)
& $d_{\text{hidden}} \cdot d_{\text{int}}$ \\

Up projection ($W_{\text{up}}$)
& $d_{\text{hidden}} \cdot d_{\text{int}}$ \\

Down projection ($W_{\text{down}}$)
& $d_{\text{int}} \cdot d_{\text{hidden}}$ \\

\noalign{\hrule height 1pt}
\textbf{FFN Total} $(\mathcal{P}_{\text{FFN}})$
& $d_\text{hidden} + 3 \cdot d_{\text{hidden}} \cdot d_{\text{int}}$ \\


\emph{Simplified} (for HAM/GDN/GDN-GSA)
& $\frac{30}{7} d^2 + d$ \\
\emph{Simplified} (for Transformer)
& $d(4d+1)$ \\
\hline

\end{tabular*}
\end{small}

\end{table}

\label{subsec:app-flops}




\begin{table}[ht]
\caption{RNN block (GDN) FLOPs.}
\label{tab:flops_rnn}
\centering
\renewcommand{\arraystretch}{1.2}

\newcommand{\colgap}{\hspace{32pt}}

\begin{small}
\begin{tabular*}{\textwidth}{l @{\extracolsep{\fill}} r}
\hline
\textbf{Operation} & \textbf{FLOPs} \\
\hline

\multicolumn{2}{l}{\textbf{RNN scalars}} \\

$\beta$ (from $b$ proj)
& $3 T d_{\mathrm{hidden}} h_{\mathrm{RNN}}$ \\

$g$ (from $a$ proj \& $A_{log}$)
& $5 T d_{\mathrm{hidden}} h_{\mathrm{RNN}}$ \\ 





\emph{Subtotal}
& $8 T d_{\mathrm{hidden}} h_{\mathrm{RNN}}$ \\
\hline

\multicolumn{2}{l}{\textbf{Chunked delta rule}} \\

$g$ cumsum
& $h_\mathrm{RNN} T$ \\

Compute $A$
& $(T/C) C^2 (2 d_\mathrm{QK} + 3.5 h_\mathrm{RNN})$ \\

Invert $A$
& $(T/C) h_\mathrm{RNN} C^3 / 2 $ \\

Recompute $w, u$
& $2 (T/C) C^2 (d_\mathrm{QK} + d_\mathrm{V})$ \\

Gated Delta Rule
& $4 T d_\mathrm{QK} d_\mathrm{RNN\text{-}V}^\mathrm{head}$ \\

Output
& $2 T (d_\mathrm{QK} d_\mathrm{RNN\text{-}V}^\mathrm{head} + C (d_\mathrm{QK} + d_\mathrm{V}))$ \\

%
%
%
%

\emph{Subtotal}
& $T \left[ h_\mathrm{RNN}(1 + 3.5C + C^2/2) + 6C d_\mathrm{QK} + 4C d_\mathrm{V} + 6 d_\mathrm{QK} d_\mathrm{RNN\text{-}V}^\mathrm{head} \right]$ \\
\noalign{\hrule height 1pt}

%
%
%
%
%
%
%


\textbf{RNN Total} $(\mathcal{F}_{\mathrm{RNN}})$
& $T \left[ h_\mathrm{RNN}(8 d_\mathrm{hidden} + 1 + 3.5C + C^2/2) + 6C d_\mathrm{QK} + 4C d_\mathrm{V} + 6 d_\mathrm{QK} d_\mathrm{RNN\text{-}V}^\mathrm{head} \right]$ \\
\hline

\end{tabular*}
\end{small}

\end{table}


\begin{table}[ht]
\caption{KV block (attention) FLOPs.}
\label{tab:flops_kv}
\centering
\renewcommand{\arraystretch}{1.2}

\newcommand{\colgap}{\hspace{32pt}}

\begin{small}
\begin{tabular*}{\textwidth}{l @{\extracolsep{\fill}} r}
\hline
\textbf{Operation} & \textbf{FLOPs} \\
\hline

\multicolumn{2}{l}{\textbf{Rotary positional encoding}} \\

RoPE on Q, K
& $4 T d_{\mathrm{QK}}$ \\

\emph{Subtotal}
& $4 T d_{\mathrm{QK}}$ \\
\hline

\multicolumn{2}{l}{\textbf{Attention}} \\

$Q K^\top$ (causal)
& $T T_{\mathrm{KV}} d_{\mathrm{QK}}$ \\


Softmax (causal)
& $2 T T_{\mathrm{KV}} h_{\mathrm{KV}}$ \\

$\mathrm{Attn} \cdot V$ (causal)
& $T T_{\mathrm{KV}} d_{\mathrm{V}}$ \\

\emph{Subtotal}
& $T T_{\mathrm{KV}} \big(
d_{\mathrm{QK}} + d_{\mathrm{V}} + 2 h_{\mathrm{KV}}
\big)$ \\
\noalign{\hrule height 1pt}

\textbf{KV Total} $(\mathcal{F}_{\mathrm{KV}})$
& $4 T d_{\mathrm{QK}}
+ T T_{\mathrm{KV}} \big(
d_{\mathrm{QK}} + d_{\mathrm{V}} + 2 h_{\mathrm{KV}}
\big)$ \\
\hline

\end{tabular*}
\end{small}

\end{table}


\begin{table*}[ht]
\caption{Full HAM layer FLOPs. References $\mathcal{F}_{\mathrm{RNN}}$ from \cref{tab:flops_rnn} and $\mathcal{F}_{\mathrm{KV}}$ from \cref{tab:flops_kv}.}
\label{tab:flops_ham}
\centering
\renewcommand{\arraystretch}{1.2}

\newcommand{\colgap}{\hspace{64pt}}

\begin{small}
\begin{tabular*}{\textwidth}{l @{\extracolsep{\fill}} r}
\hline
\textbf{Operation} & \textbf{FLOPs} \\
\hline

\multicolumn{2}{l}{\textbf{Pre-norm}} \\
Pre-norm
& $4 T d_{\mathrm{hidden}}$ \\

\emph{Subtotal}
& $4 T d_{\mathrm{hidden}}$ \\
\hline

\multicolumn{2}{l}{\textbf{Q, K, V projections} (shared)} \\

%

%

$W_{\mathrm{Q}}, W_{\mathrm{K}}$
& $4 T d_{\mathrm{hidden}} d_{\mathrm{QK}}$ \\

$W_{\mathrm{V}}$
& $2 T d_{\mathrm{hidden}} d_{\mathrm{V}}$ \\

\emph{Subtotal}
& $2 T d_{\mathrm{hidden}} (2 d_{\mathrm{QK}} + d_{\mathrm{V}})$ \\
\hline

\multicolumn{2}{l}{\textbf{RMS norms for Q, K, V} (separate)} \\

RNN Q, K, V norms
& $4 T (2 d_{\mathrm{QK}} + d_{\mathrm{V}})$ \\

KV Q, K, V norms
& $4 T (2 d_{\mathrm{QK}} + d_{\mathrm{V}})$ \\

\emph{Subtotal}
& $8 T \big( 2 d_{\mathrm{QK}} + d_{\mathrm{V}} \big)$ \\
\hline

\multicolumn{2}{l}{\textbf{1D convolutions} (separate)} \\

RNN Q, K, V conv
& $T (2 d_{\mathrm{QK}} + d_{\mathrm{V}}) (2 d_{\mathrm{conv}} + 3)$ \\

KV Q, K, V conv
& $T (2 d_{\mathrm{QK}} + d_{\mathrm{V}}) (2 d_{\mathrm{conv}} + 3)$ \\

\emph{Subtotal}
& $2 T (2 d_{\mathrm{QK}} + d_{\mathrm{V}}) (2 d_{\mathrm{conv}} + 3)$ \\
\hline

\multicolumn{2}{l}{\textbf{KV token selection}} \\

via RNN state (if not using a learnable router)
& $12 T d_{\mathrm{hidden}}$ \\

via input (if using a learnable router)
& $2 T d_\mathrm{hidden}$ \\

\emph{Subtotal}
& $2 T d_{\mathrm{hidden}} + 10 T d_{\mathrm{hidden}} \mathbbm{1}(\neg \text{learnable router})$ \\
\hline

\multicolumn{2}{l}{\textbf{RNN block}} \\

RNN forward (\cref{tab:flops_rnn})
& $\mathcal{F}_{\mathrm{RNN}}$ \\

\emph{Subtotal}
& $T \left[ h_\mathrm{RNN}(8 d_\mathrm{hidden} + 1 + 3.5C + C^2/2) + 6C d_\mathrm{QK} + 4C d_\mathrm{V} + 6 d_\mathrm{QK} d_\mathrm{RNN\text{-}V}^\mathrm{head} \right]$ \\
\hline

\multicolumn{2}{l}{\textbf{KV block}} \\

KV forward (\cref{tab:flops_kv})
& $\mathcal{F}_{\mathrm{KV}}$ \\

\emph{Subtotal}
& $4 T d_{\mathrm{QK}} + T T_{\mathrm{KV}} ( d_{\mathrm{QK}} + d_{\mathrm{V}} + 2 h_{\mathrm{KV}} )$ \\
\hline

\multicolumn{2}{l}{\textbf{RNN \& KV headwise output normalization}} \\

RNN norm
& $4 T d_{\mathrm{RNN\text{-}V}}^{\mathrm{head}}$ \\

RNN norm gate
& $2 T d_{\mathrm{hidden}} d_{\mathrm{V}} + T d_{\mathrm{V}}$ \\

KV norm
& $4 T d_{\mathrm{KV\text{-}V}}^{\mathrm{head}}$ \\

\emph{Subtotal}
& $4 T (d_{\mathrm{RNN\text{-}V}}^{\mathrm{head}} + d_{\mathrm{KV\text{-}V}}^{\mathrm{head}}) + 2 T d_{\mathrm{hidden}} d_{\mathrm{V}} + T d_{\mathrm{V}}$ \\
\hline

\multicolumn{2}{l}{\textbf{RNN \& KV gating}} \\

Per-head RNN gate
& $2 T d_{\mathrm{hidden}} h_{\mathrm{RNN}} + T d_{\mathrm{V}}$ \\

Per-head KV gate
& $2 T d_{\mathrm{hidden}} h_{\mathrm{KV}} + T d_{\mathrm{V}}$ \\

\emph{Subtotal}
& $2 T d_{\mathrm{hidden}} (h_{\mathrm{RNN}} + h_{\mathrm{KV}}) + 2 T d_{\mathrm{V}}$ \\
\hline

%
%
%
%
%

\multicolumn{2}{l}{\textbf{Output projection}} \\

Output projection
& $2 T d_{\mathrm{V}} d_{\mathrm{hidden}}$ \\

\emph{Subtotal}
& $2 T d_{\mathrm{V}} d_{\mathrm{hidden}}$ \\

\noalign{\hrule height 1pt}
\textbf{Layer Total}
& $T \Big[ 4 d_{\mathrm{hidden}} d_{\mathrm{QK}} + 6 d_{\mathrm{hidden}} d_{\mathrm{V}} + 8 d_{\mathrm{conv}} d_{\mathrm{QK}} + 4 d_{\mathrm{conv}} d_{\mathrm{V}} + 6 d_{\mathrm{QK}} d_{\mathrm{RNN\text{-}V}}^{\mathrm{head}}$ \\
& $\quad + \; d_{\mathrm{hidden}} (16 + 2 h_{\mathrm{KV}}) + d_{\mathrm{QK}} (32 + 6C + T_{\mathrm{KV}}) + d_{\mathrm{V}} (17 + 4C + T_{\mathrm{KV}})$ \\
& $\quad + \; h_{\mathrm{RNN}} (10 d_{\mathrm{hidden}} + 1 + 3.5C + C^2/2) + 2 T_{\mathrm{KV}} h_{\mathrm{KV}} + 4 d_{\mathrm{RNN\text{-}V}}^{\mathrm{head}} + 4 d_{\mathrm{KV\text{-}V}}^{\mathrm{head}} \Big]$ \\



\emph{Simplified}
& $T\left[ \frac{65}{7}d^2 + \frac{33059}{14}d +13669 + T_{\mathrm{KV}}\left(\frac{25}{14}d + 20\right) \right]$ \\
\hline

\end{tabular*}
\end{small}

\end{table*}

\begin{table*}[ht]
\caption{Transformer attention layer FLOPs (excluding FFN).}
\label{tab:flops_transformer}
\centering
\renewcommand{\arraystretch}{1.2}

\newcommand{\colgap}{\hspace{40pt}}

\begin{small}
\begin{tabular*}{\textwidth}{l @{\extracolsep{\fill}} r}
\hline
\textbf{Operation} & \textbf{FLOPs} \\
\hline

\multicolumn{2}{l}{\textbf{Pre-norm}} \\

Pre-norm
& $4 T d_{\mathrm{hidden}}$ \\

\emph{Subtotal}
& $4 T d_{\mathrm{hidden}}$ \\
\hline

\multicolumn{2}{l}{\textbf{Q, K, V projections}} \\

$W_Q$
& $2 T d_{\mathrm{hidden}}^2$ \\

$W_K$
& $2 T d_{\mathrm{hidden}} d_{\mathrm{KV}}$ \\

$W_V$
& $2 T d_{\mathrm{hidden}} d_{\mathrm{KV}}$ \\

\emph{Subtotal}
& $2 T d_{\mathrm{hidden}} (d_{\mathrm{hidden}} + 2 d_{\mathrm{KV}})$ \\
\hline

\multicolumn{2}{l}{\textbf{Rotary positional encoding}} \\

RoPE on Q
& $6 T d_{\mathrm{hidden}}$ \\

RoPE on K
& $6 T d_{\mathrm{KV}}$ \\

\emph{Subtotal}
& $6 T (d_{\mathrm{hidden}} + d_{\mathrm{KV}})$ \\
\hline

\multicolumn{2}{l}{\textbf{Attention}} \\

$Q K^\top$ (causal)
& $T^2 d_{\mathrm{hidden}}$ \\


Softmax (causal)
& $2 T^2 h$ \\

$\mathrm{Attn} \cdot V$ (causal)
& $T^2 d_{\mathrm{hidden}}$ \\

\emph{Subtotal}
& $2 T^2 (d_{\mathrm{hidden}} + h)$ \\
\hline

\multicolumn{2}{l}{\textbf{Output projection}} \\

$W_O$
& $2 T d_{\mathrm{hidden}}^2$ \\

\emph{Subtotal}
& $2 T d_{\mathrm{hidden}}^2$ \\
\noalign{\hrule height 1pt}

\textbf{Layer Total}
& $2 T [ 2 d_{\mathrm{hidden}} + d_{\mathrm{hidden}} (d_{\mathrm{hidden}} + 2 d_{\mathrm{KV}}) + 3 (d_{\mathrm{hidden}} + d_{\mathrm{KV}}) + d_{\mathrm{hidden}}^2 ] + 2 T^2 (d_{\mathrm{hidden}} + h)$ \\




\emph{Simplified}
& $8Td (d+2) + \frac{129}{64} T^2 d$ \\

\hline

\end{tabular*}
\end{small}

\end{table*}

\begin{table*}[ht]
\caption{Pure Gated DeltaNet layer FLOPs.}
\label{tab:flops_gdn}
\centering
\renewcommand{\arraystretch}{1.2}

\newcommand{\colgap}{\hspace{64pt}}

\begin{small}
\begin{tabular*}{\textwidth}{l @{\extracolsep{\fill}} r}
\hline
\textbf{Operation} & \textbf{FLOPs} \\
\hline

\multicolumn{2}{l}{\textbf{Pre-norm}} \\
Pre-norm
& $4 T d_{\mathrm{hidden}}$ \\

\emph{Subtotal}
& $4 T d_{\mathrm{hidden}}$ \\
\hline

\multicolumn{2}{l}{\textbf{Q, K, V projections}} \\

$W_{\text{Q}}, W_{\text{K}}$
& $4 T d_{\text{hidden}} d_{\mathrm{QK}}$ \\

$W_{\mathrm{V}}$
& $2 T d_{\text{hidden}} d_{\mathrm{V}}$ \\

\emph{Subtotal}
& $2 T d_{\text{hidden}} (2 d_{\mathrm{QK}} + d_{\mathrm{V}})$ \\
\hline

\multicolumn{2}{l}{\textbf{RMS norms for Q, K, V}} \\

Q, K, V norms
& $4 T (2 d_{\mathrm{QK}} + d_{\mathrm{V}})$ \\

\emph{Subtotal}
& $4 T (2 d_{\mathrm{QK}} + d_{\mathrm{V}})$ \\
\hline

\multicolumn{2}{l}{\textbf{1D convolutions}} \\

Q, K, V conv
& $T (2 d_{\mathrm{QK}} + d_{\mathrm{V}}) (2 d_{\text{conv}} + 3)$ \\

\emph{Subtotal}
& $(2 d_{\text{conv}} + 3) T (2 d_{\mathrm{QK}} + d_{\mathrm{V}})$ \\
\hline

\multicolumn{2}{l}{\textbf{RNN scalars}} \\

$\beta$ (from $b$ proj)
& $3 T d_{\text{hidden}} h$ \\

$g$ (from $a$ proj \& $A_{\log}$)
& $5 T d_{\text{hidden}} h$ \\

\emph{Subtotal}
& $8 T d_{\text{hidden}} h$ \\
\hline

\multicolumn{2}{l}{\textbf{Chunked delta rule}} \\

$g$ cumsum
& $h T$ \\

Compute $A$
& $(T/C) C^2 (2 d_{\mathrm{QK}} + 3.5 h)$ \\

Invert $A$
& $(T/C) h C^3 / 2$ \\

Recompute $w, u$
& $2 (T/C) C^2 (d_{\mathrm{QK}} + d_{\mathrm{V}})$ \\

Gated delta rule
& $4 T d_{\mathrm{QK}} d_{\mathrm{V}}^{\text{head}}$ \\

Output
& $2 T (d_{\mathrm{QK}} d_{\mathrm{V}}^{\text{head}} + C (d_{\mathrm{QK}} + d_{\mathrm{V}}))$ \\

\emph{Subtotal}
& $T \left[ h(1 + 3.5C + C^2/2) + 6C \cdot d_{\mathrm{QK}} + 4C \cdot d_{\mathrm{V}} + 6 \cdot d_{\mathrm{QK}} \cdot d_{\mathrm{V}}^{\text{head}} \right]$ \\
\hline

%
%

\multicolumn{2}{l}{\textbf{Output gating}} \\

Pre-gate norm
& $4 T d_{\text{hidden}}$ \\

Gate projection
& $2 T d_{\text{hidden}} d_{\mathrm{V}}$ \\

SiLU activation
& $3 T d_{\mathrm{V}}$ \\

Fused norm + gate
& $5 T d_{\mathrm{V}}$ \\

\emph{Subtotal}
& $T (4 d_{\text{hidden}} + 2 d_{\text{hidden}} d_{\mathrm{V}} + 8 d_{\mathrm{V}})$ \\
\hline

\multicolumn{2}{l}{\textbf{Output projection}} \\

Output projection
& $2 T d_{\mathrm{V}} d_{\text{hidden}}$ \\

Final norm
& $4 T d_{\text{hidden}}$ \\

\emph{Subtotal}
& $2 T d_{\text{hidden}} (d_{\mathrm{V}} + 2)$ \\

\noalign{\hrule height 1pt}
\textbf{Layer Total}
& $T \big[ (2 d_{\text{hidden}} + 2 d_{\text{conv}} + 7)(2 d_{\mathrm{QK}} + d_{\mathrm{V}}) + 8 d_{\text{hidden}} h + h(1 + 3.5C + C^2/2)$ \\
& $\quad + 6C d_{\mathrm{QK}} + 4C d_{\mathrm{V}} + 6 d_{\mathrm{QK}} d_{\mathrm{V}}^{\text{head}} + 12 d_{\text{hidden}} + 4 d_{\text{hidden}} d_{\mathrm{V}} + 8 d_{\mathrm{V}} \big]$ \\


\emph{Simplified}
& $T d \left[\frac{2085}{224}d+\frac{1560037}{1792}\right]$ \\
\hline

\end{tabular*}
\end{small}

\end{table*}

\begin{table}[ht]
  \caption{FFN (SwiGLU) FLOPs.}
  \label{tab:flops_ffn}
  \centering
  \renewcommand{\arraystretch}{1.2}

  \newcommand{\colgap}{\hspace{32pt}}

  \begin{small}
  \begin{tabular*}{\textwidth}{l @{\extracolsep{\fill}} r}
  \hline
  \textbf{Operation} & \textbf{FLOPs} \\
  \hline

  \multicolumn{2}{l}{\textbf{FFN (SwiGLU)}} \\

  Gate projection ($W_{\text{gate}}$)
  & $2 T d_{\text{hidden}} d_{\text{int}}$ \\

  Up projection ($W_{\text{up}}$)
  & $2 T d_{\text{hidden}} d_{\text{int}}$ \\

  SiLU activation
  & $3 T d_{\text{int}}$ \\

  Gate $\cdot$ Up
  & $T d_{\text{int}}$ \\

  Down projection ($W_{\text{down}}$)
  & $2 T d_{\text{int}} d_{\text{hidden}}$ \\

  \hline

  \textbf{Layer Total}
  & $T d_{\text{int}} (6 d_{\text{hidden}} + 4)$ \\


\emph{Simplified} (for HAM/GDN/GDN-GSA)
  & $\frac{20}{7} T d (3d + 2)$ \\
\emph{Simplified} (for Transformer)
  & $\frac{8}{3} T d (3d + 2)$ \\
  \hline

  \end{tabular*}
  \end{small}

\end{table}

  \end{document}